\documentclass[10pt,compsoc]{IEEEtran}
\ifCLASSOPTIONcompsoc
  \usepackage[nocompress]{cite}
\else
  \usepackage{cite}
\fi
\ifCLASSINFOpdf
\else
\fi
\usepackage{epstopdf}
\usepackage{algorithm}
\usepackage{algorithmic}
\usepackage{url}
\usepackage{color}
\usepackage{adjustbox}
\usepackage{rotating}
\usepackage{lineno,hyperref}
\usepackage{amsmath}
\usepackage{amssymb}
\usepackage{subfigure}
\usepackage{balance}
\usepackage{booktabs}
\modulolinenumbers[5]
\usepackage{bbding}
\usepackage{multirow} 
\usepackage{multicol} 
\graphicspath{ {image/} }

\usepackage{lineno,hyperref}

\newtheorem{myDef}{Definition}

\begin{document}
%
\title{A Comprehensive Survey on \\Source-free Domain Adaptation}
%
%
%
%

\author{Zhiqi Yu, Jingjing Li, Zhekai Du, Lei Zhu, Heng Tao Shen
\thanks{The authors are with the School of Computer Science and Engineering, University of Electronic Science and Technology of China, Chengdu, 611731.}}

%
%

\markboth{A Comprehensive Survey on Source-free Domain Adaptation}%
{Yu \MakeLowercase{\textit{et al.}}: A Comprehensive Survey on Source-free Domain Adaptation}
%



\IEEEtitleabstractindextext{%
\begin{abstract}
Over the past decade, domain adaptation has become a widely studied branch of transfer learning that aims to improve performance on target domains by leveraging knowledge from the source domain. Conventional domain adaptation methods often assume access to both source and target domain data simultaneously, which may not be feasible in real-world scenarios due to privacy and confidentiality concerns. As a result, the research of Source-Free Domain Adaptation (SFDA) has drawn growing attention in recent years, which only utilizes the source-trained model and unlabeled target data to adapt to the target domain. Despite the rapid explosion of SFDA work, yet there has no timely and comprehensive survey in the field. To fill this gap, we provide a comprehensive survey of recent advances in SFDA and organize them into a unified categorization scheme based on the framework of transfer learning. Instead of presenting each approach independently, we modularize several components of each method to more clearly illustrate their relationships and mechanics in light of the composite properties of each method. Furthermore, we compare the results of more than 30 representative SFDA methods on three popular classification benchmarks, namely Office-31, Office-home, and VisDA, to explore the effectiveness of various technical routes and the combination effects among them. Additionally, we briefly introduce the applications of SFDA and related fields. Drawing from our analysis of the challenges facing SFDA, we offer some insights into future research directions and potential settings.


\end{abstract}

\begin{IEEEkeywords}
Domain Adaptation, Transfer Learning, Computer Vision, Data-Free Learning.
\end{IEEEkeywords}}

\maketitle

\IEEEdisplaynontitleabstractindextext

%
\IEEEpeerreviewmaketitle

\IEEEraisesectionheading{\section{Introduction}\label{sec:introduction}}
Deep neural networks are able to achieve satisfying performance in supervised learning tasks thanks to their generalization ability. However, collecting sufficient training data can be expensive due to factors such as cost and privacy concerns, etc. For instance, manually annotating a single cityscape image for semantic segmentation can take up to 90 minutes. To alleviate this issue, transfer learning~\cite{perkins1992transfer,pan2009survey} has been proposed to enable cross-domain knowledge transfer under label deficient conditions. 

Domain adaptation (DA) \cite{li2020maximum,jing2022adversarial}, as an important branch of transfer learning, focuses on how to improve model performance on unlabeled target domains with the assistance of labeled source domain data. Under the condition of independent homogeneous distribution, the biggest challenge of domain adaptation is how to reduce the domain shift, which can be mainly classified as conditional shift, covariate shift, label shift and concept shift, and has been widely discussed in the previous DA surveys~\cite{wilson2020survey,liu2022deep}.

In conventional domain adaptation approaches, it is mostly assumed that both the source and target domain data could be accessed during adaptation. However, this is not realistic in many applications. On the one hand, the raw source domain data will be unavailable in some cases due to personal privacy, confidentiality and copyright issues; on the other hand, it is unrealistic to keep the complete source dataset on some resource-limited devices for training. All the above problems have hindered the further spread of this field. To relax the dependence of DA methods on source data, Source-Free Domain Adaption (SFDA)~\cite{liang2020we,chidlovskii2016domain} has been proposed in recent years, which has rapidly become the focus of domain adaptation and is widely studied in image classification~\cite{kundu2020universal,xia2021adaptive}, semantic segmentation~\cite{liu2021source,bateson2022source} and object detection~\cite{xiong2022source,saltori2020sf}.

\begin{figure}[t]
               \centering
               \subfigure[UDA]{
               \includegraphics[width=0.45\linewidth] {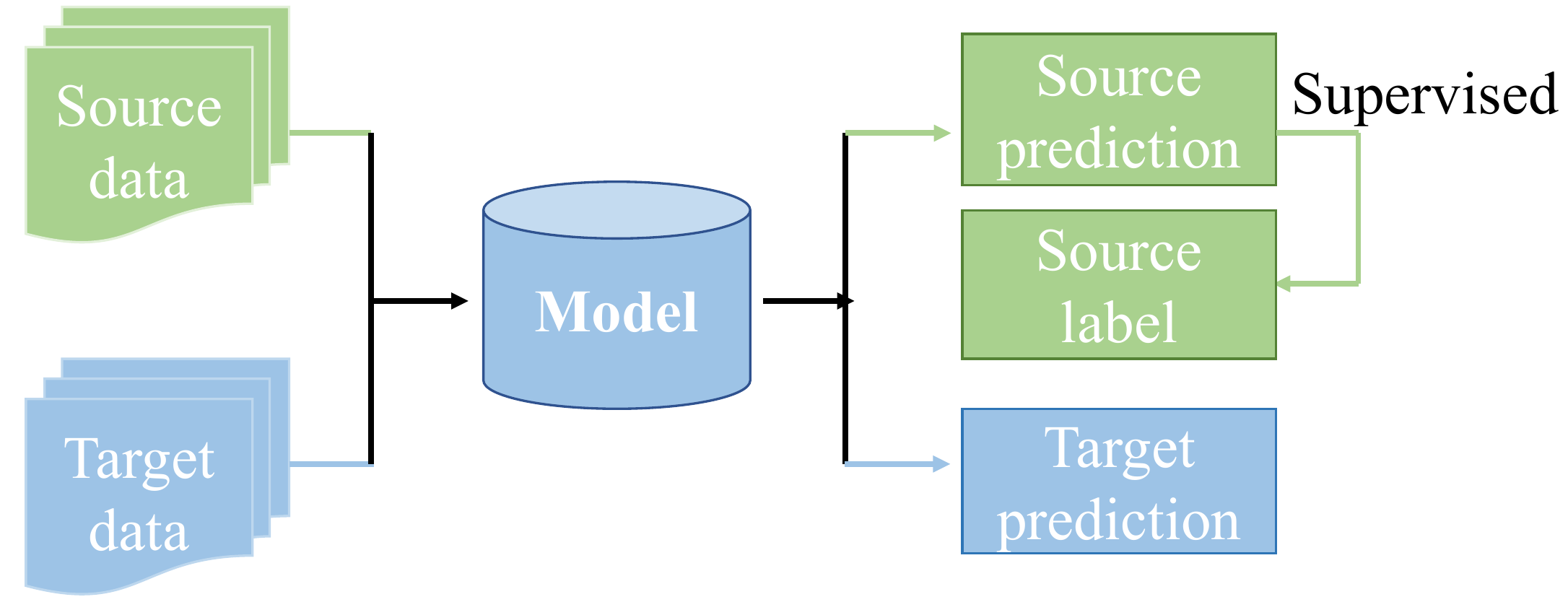}
               \label{fig:parameter}
               }
               \subfigure[SFDA]{
               \includegraphics[width=0.45\linewidth] {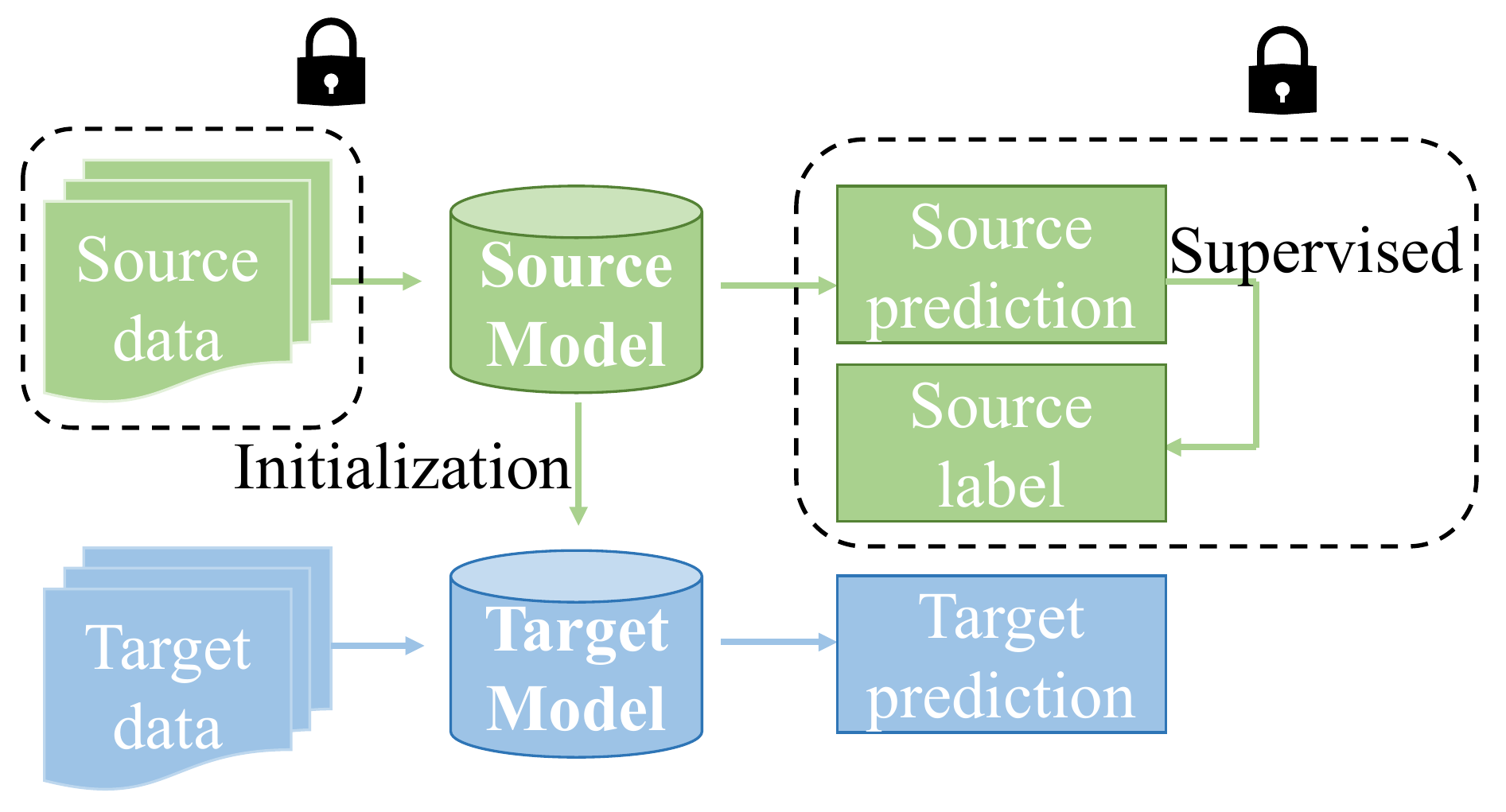}
               \label{fig:ablation}
               }
               \label{myfigure}
                \vspace{-7pt}
               \caption{Comparison of UDA and SFDA settings.}
               \vspace{-10pt}
               \label{fig:UDA_SFDA}
           \end{figure}

\begin{figure*}[t]
\centering
\includegraphics[width=0.8\linewidth]{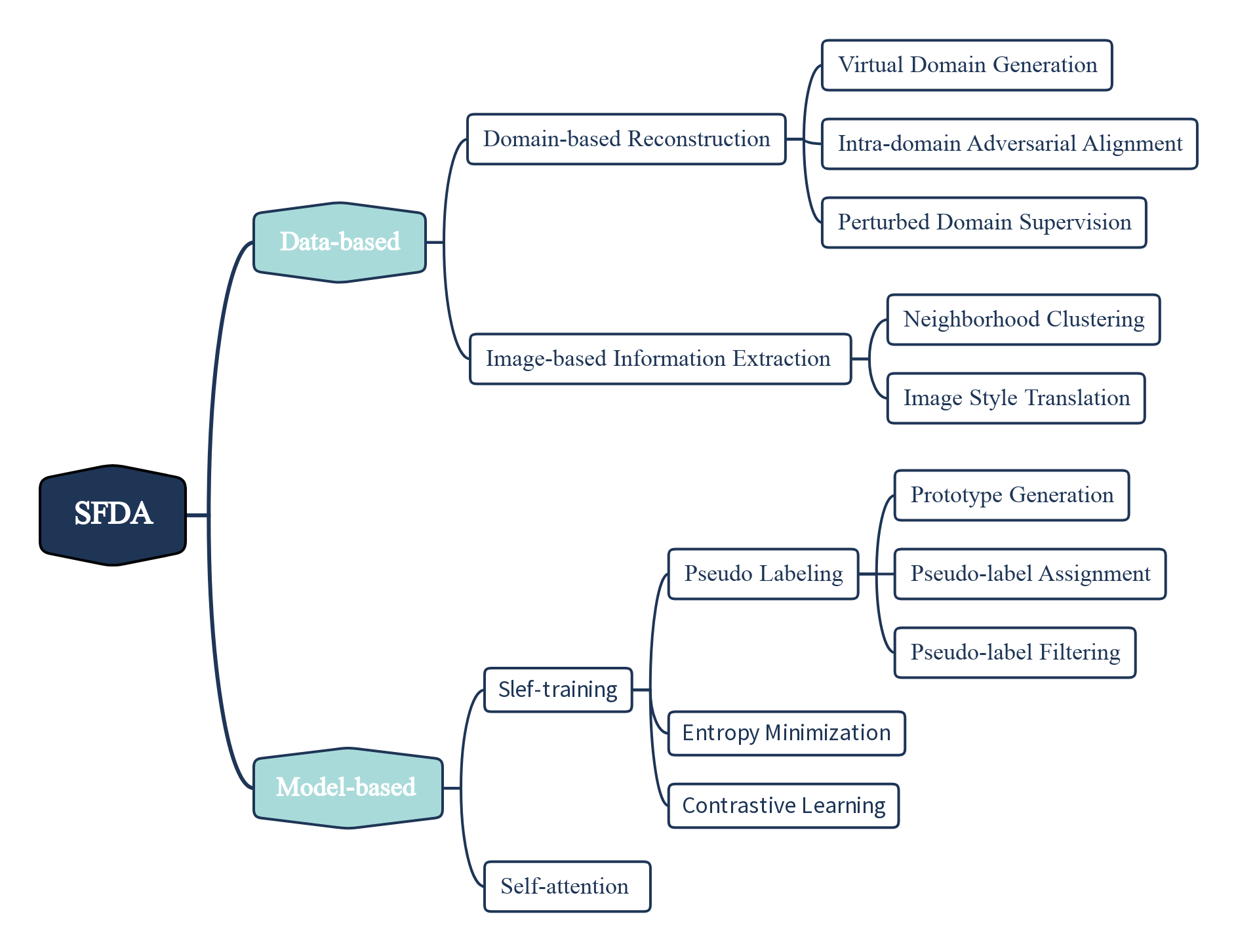}
\vspace{-5pt}
\caption{Taxonomy of SFDA methods.}
\label{fig:framework}
\vspace{-10pt}
\end{figure*}

 As shown in Fig.~\ref{fig:UDA_SFDA}, the most significant difference between SFDA and Unsupervised Domain Adaptation (UDA) \cite{ben2010theory, long2016unsupervised} is that the UDA model can be trained using both the source and target domain data, while SFDA can only utilize the source model to initialize the target model and then update it with the unlabeled target data. Existing mainstream UDA methods can be categorized into two main types of methods: those that align the source and target domain distributions by designing specific metrics \cite{li2020maximum,long2015learning,cui2020towards}, and those that learn domain-invariant feature representations through adversarial learning \cite{ganin2015unsupervised,long2018conditional,tzeng2017adversarial}. However, these mainstream UDA methods are not suitable for scenarios without source domain data, highlighting the need to investigate SFDA methods as an alternative. Despite the considerable efforts devoted to SFDA, there has been no comprehensive review of all the works and a summary of the current progress related to SFDA. To fill this gap, we aim to provide a timely and comprehensive review of recent advances in SFDA in this survey.

In this work, we inherit from existing transfer learning surveys~\cite{pan2009survey,zhuang2020comprehensive} and divide SFDA into two directions: data-based one and model-based one. In addition, we find that conventional UDA methods begin to work in the SFDA setting through some additional steps with the reconstructed source domain, and thus we discuss this regime as the domain-based reconstruction methods to reflect the extensibility of UDA research. We depict the overall topology of SFDA methods in Fig.~\ref{fig:framework}. 

From the data-centric perspective, the domain-based reconstruction and the image style translation can be seen as some derivatives of UDA. The intuition behind domain-based reconstruction is quite straightforward, i.e., it aims to reconstruct a domain or make further divisions within the target domain to compensate for missing source domain data, so that the UDA approaches can be extended to the SFDA setting. Concretely, it can be subdivided into virtual domain generation, intra-domain adversarial alignment and perturbed domain supervision. Image style translation translates the target domain data into the unseen source-style through the Batch Normalization (BN) layers of the source classifier, thereby the target domain data can be better compatible with the source model. Neighborhood clustering is based on the observation that the underlying intrinsic structure in the target data is embedded in the neighbor relationship, even though the target domain data distribution can not explicitly align with the source classifier. One assumption is that each category of the target domain will exist neighbors located within the source model's boundary, so those hard samples can be mapped to the source domain distribution by maintaining consistency among neighbors. At last, local structure clustering is based on manifold learning \cite{huang2015new,li2018generalized,han2001text}, and clustering is performed through the selection of neighbor nodes.

From the model-centric perspective, the majority of approaches assume that the source-pretrained model has a certain degree of generalization over the target domain due to the similarity of the source and target domains. Therefore, the model can be fine-tuned by exploring the outputs of the model on the target data. This self-training scheme is inherited from the idea of semi-supervised learning \cite{saito2019semi,kumar2010co,li2018semi}, and can be futher divided into pseudo-labeling, entropy minimization and contrastive learning. It is noted that we pay additional attention to the pseudo-labeling methods since it is most commonly used in SFDA. Pseudo-labeling is usually achieved by first pseudo-labeling the high-confidence samples in the target domain, and optimize the model thereafter by the obtained pseudo-labels. Therefore, how to obtain high-quality pseudo-label is the main concern of this category of methods. In this paper we try to discuss this technical route comprehensively based on each process in terms of prototype generation, pseudo-label assignment and pseudo-label filtering.

\begin{table*}[t]
\renewcommand\arraystretch{1.5} 
\centering
\caption{SFDA versus other related settings.}
\vspace{-10pt}
\begin{tabular}{c|c|c|c|c|c}
\toprule
               &No source data &  Source model & Unlabeled target & Extra source information & No training\\
\hline
Data-free KD           &    \CheckmarkBold            &      -     &      -   &    -     & \XSolidBrush\\ 
Federated DA            &  \XSolidBrush    &      \CheckmarkBold      &       \CheckmarkBold    &    \XSolidBrush  & \XSolidBrush\\   
Zero-shot DA           &  \XSolidBrush    &       \CheckmarkBold     &       \XSolidBrush   &   \XSolidBrush  & \XSolidBrush\\    
Test-time DA           &  \CheckmarkBold    &      \CheckmarkBold    &       \CheckmarkBold    &  \XSolidBrush   &    \CheckmarkBold\\    
Source-relaxed DA      &    \CheckmarkBold   &      \CheckmarkBold      &        \CheckmarkBold    &   \CheckmarkBold   & \XSolidBrush\\    
\hline
Source-free DA           &     \CheckmarkBold  &       \CheckmarkBold     &     \CheckmarkBold  &    \XSolidBrush    &   \XSolidBrush\\
\hline
\end{tabular}
\label{tab:setting}
\end{table*}

Moreover, we observed that most of the SFDA methods are comprised of multiple technical components, and each component might correspond to a category in our taxonomy. Therefore, unlike previous surveys on domain adaptation that treated each work independently, we aim to modularize each SFDA method and categorize them to reveal the interrelatedness among different research directions. The goal of this survey is to discuss the problem formulation of SFDA, summarize the current advances in this field, and highlight the distinctive components of SFDA techniques to provide a comprehensive understanding of this area and inspire more ideas and applications. The main contributions of this survey can be summarized as follows.

\begin{itemize}
\item[$\bullet$] We propose an overall taxonomy framework for the newly emerged SFDA setting from both data-centric and model-centric perspectives. This survey fills a gap in the existing literature by providing a timely and comprehensive summary of the latest SFDA methods and applications.
\end{itemize}

\begin{itemize}
\item[$\bullet$] We modularize more than 30 representative works and compare the results on the most widely used classification datasets, i.e., Office-31, Office-home and VisDA. The combination of different components is visually presented and comparatively analyzed, which may be instructive for further research in the SFDA setting.
\end{itemize}

The remainder of this survey is structured as follows. In Section 2, we introduce the notations and preliminary knowledge related to SFDA. Section 3 and Section 4 provide a comprehensive review of data-centric and model-centric SFDA methods, respectively. Section 5 presents a comparison of existing SFDA methods on three mainstream classification datasets. Section 6 discusses the potential applications of SFDA, while Section 7 presents visions for future directions of SFDA. Finally, we conclude the paper in Section 8.

\section{Overview}

In this section, we introduce some notations and the preliminary knowledge related to SFDA, with a brief description of several settings in SFDA methods.

\subsection{Notations and Definitions} 
For consistency among the literature, we used as many notations as possible that are the same as those in the surveys related to transfer learning~\cite{zhuang2020comprehensive} and deep domain adaptation~\cite{liu2022deep,wang2018deep}. We summarize them in Table~\ref{tab:notation}. Below are some basic definitions that are closely related with the problem of SFDA.

\begin{table}[h]
\renewcommand\arraystretch{1.5} 
\centering
\caption{Notations related to SFDA.}
\vspace{-10pt}
\begin{tabular}{c|c}
\toprule
   \textbf{Symbol}            &\textbf{Description} \\
\hline
$n$          &    Number of instances         \\ 
$\mathcal{D}$          &    Domain         \\ 
$\mathcal{X}$          &    Instance set      \\ 
$d$          &    Feature dimensions\\ 
$\mathcal{F}(\mathcal{X})$          &    Feature  vector    \\
$\mathcal{Y}|\mathcal{X}$ & Prediction for instance $\mathcal{X}$\\
$\Phi$          &   Domain space         \\ 
$\mathbf{L}$            &    Source label set         \\ 
$\mathcal{L}$            &    Loss function        \\ 
$\delta$            &    Soft-max operation       \\ 
$c$            &    Number of classes       \\ 
$\widetilde{z}$           &    Noise vector        \\
$\Theta$            & Model parameters   \\
$\mathcal{F}$       &    Feature extractor         \\
$\mathcal{C}$       &    Classifier          \\
$\mathcal{D}$       &    Domain discriminator          \\
$G$       &    Generator          \\
$\mathcal{M}$ &         Model to be trained                   \\
$\mathbf{M}$ &         Module                   \\
$\mathcal{B}$ &         Memory bank                  \\
$\mu$ &         Means                   \\
$\sigma$ &         Standard deviation                  \\
$\mathbb{E} $ &         Mathematical expectation                  \\
$P$             &       Marginal distribution    \\
$Q$             &       Query    \\
$K$             &       Key   \\
$V$             &       Value   \\
$\mathcal{N}(\mu,\sigma^{2})$ &      A Gaussian distribution with mean $\mu$ and variance $\sigma^{2}$                  \\
\hline
\end{tabular}
\label{tab:notation}
\end{table}

\begin{myDef}
    (Domain in SFDA) A domain usually contains a data set $\Phi$ and a corresponding label set $\mathbf{L}$, where the data set contains the instance set $\mathcal{X}$ drown from the marginal distribution $\mathcal{P\left(X\right)}$ and corresponding dimension $d$. In SFDA, the source domain  $\mathcal{D}^{s}=\{\{\mathcal{X}^{s},\mathcal{P}\left(\mathcal{X}^{s}\right), d^s\},\mathbf{L}^{s}\}$ is usually complete, but note that it is only available during pre-training. The label set of the target domain $\mathcal{D}^{t}=\{\{\mathcal{X}^{t},\mathcal{P}\left(\mathcal{X}^{t}\right),d^t\}\}$ is not available and its marginal distribution is also different from the source domain. In typical SFDA, we assume a closed form, i.e., the label spaces of the source and target domains are the same.
\end{myDef}

\begin{myDef}
    (Task of SFDA) Given a model $\mathcal{M}$ trained on the source domain $\mathcal{D}^{s}=\{\{\mathcal{X}^{s},\mathcal{P}\left(\mathcal{X}^{s}\right),d^s\},\mathbf{L}^{s}\}$, the process of SFDA consists of two stages. In the first stage (pre-training), the source model is trained by labeled source data under supervised learning. In the second stage (adaptation), the source domain data is unavailable, and the goal of SFDA is to adapt the source-trained model to the target domain with unlabeled target data $\Phi^{t}=\{\{\mathcal{X}^{t},\mathcal{P\left(X\right)}^{t}, d^t\}\}$, i.e., to get high accuracy for outputs $\mathcal{M}(\mathcal{X}^{t})=P(\mathcal{Y}^{t}|\mathcal{X}^{t}$), where $P\left(\mathcal{Y}^{t}|\mathcal{X}^{t}\right)$ is the prediction result of target data $\mathcal{X}^{t}$.
\end{myDef}

It is worth noting that the above definitions may also enable some extensions. For the definition of domain, SFDA may include the multi-source domains\cite{dong2021confident,ahmed2021unsupervised}, i.e., $\mathcal{D}^{s}=\{\sum_{i=1}^{m}\mathcal{D}^{s}_{i}\}$ where m is the number of source domains, or there may be non-closed set cases such as Universal source-free domain adaptation\cite{kundu2020universal}; for the definition of task in SFDA, for example, it is also possible in some extended SFDA settings to use a small portion of labeled target domain data when performing the task (i.e., Active SFDA~\cite{wang2022active,su2022comparison}). We will discuss different variants of SFDA in the next section.

\subsection{Overview of the settings in SFDA}

\begin{figure}[t]
\centering
\label{fig:intro}
\includegraphics[width=1.0\linewidth]{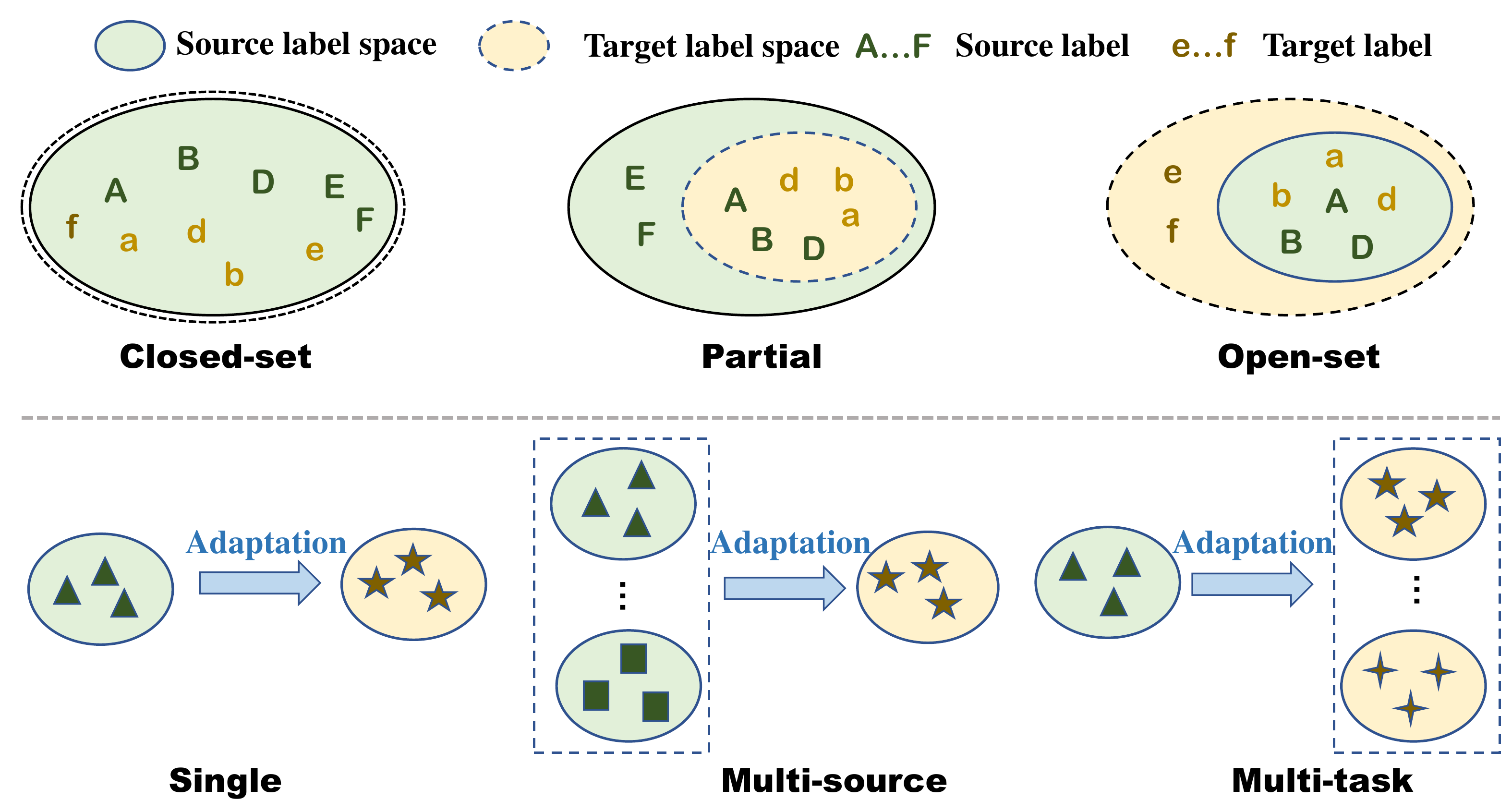}
\vspace{-5pt}
\caption{ An illustration of different settings in SFDA.}
\label{fig:label_set}
\vspace{-5pt}
\end{figure}

Generally, domain adaptation can be divided into homogeneous domain adaptation and heterogeneous domain adaptation according to whether the source and target domain feature spaces are identical. In heterogeneous domain adaptation \cite{day2017survey}, both the distribution space and feature space of the source and target may be different, i.e., $P(\mathcal{X}^s) \neq P(\mathcal{X}^t)$ and $d^s \neq d^t$, so both distribution adaptation and feature space adaptation may be required. While in homogeneous domain adaptation, the feature spaces of the source and target domains are identical and only distribution adaptation is performed. Among the existing SFDA methods, most of them only involve homogeneous domain adaptation, which is also the focus of our discussion in this survey.

According to the label-set correspondence between the source and target domains, the settings can be classified as closed-set, partial, and open-set, and the relationship between them are shown in Fig.~\ref{fig:label_set}. Based on the number of source and target domains, there are also single-source, multi-source, and multi-task settings. In this survey, we mainly focus on the most general closed-set single-source SFDA setting, and other settings are also briefly outlined.

\begin{figure*}[h]
              \centering
              \subfigure[Virtual Domain Generation]{
              \includegraphics[width=0.27\linewidth] {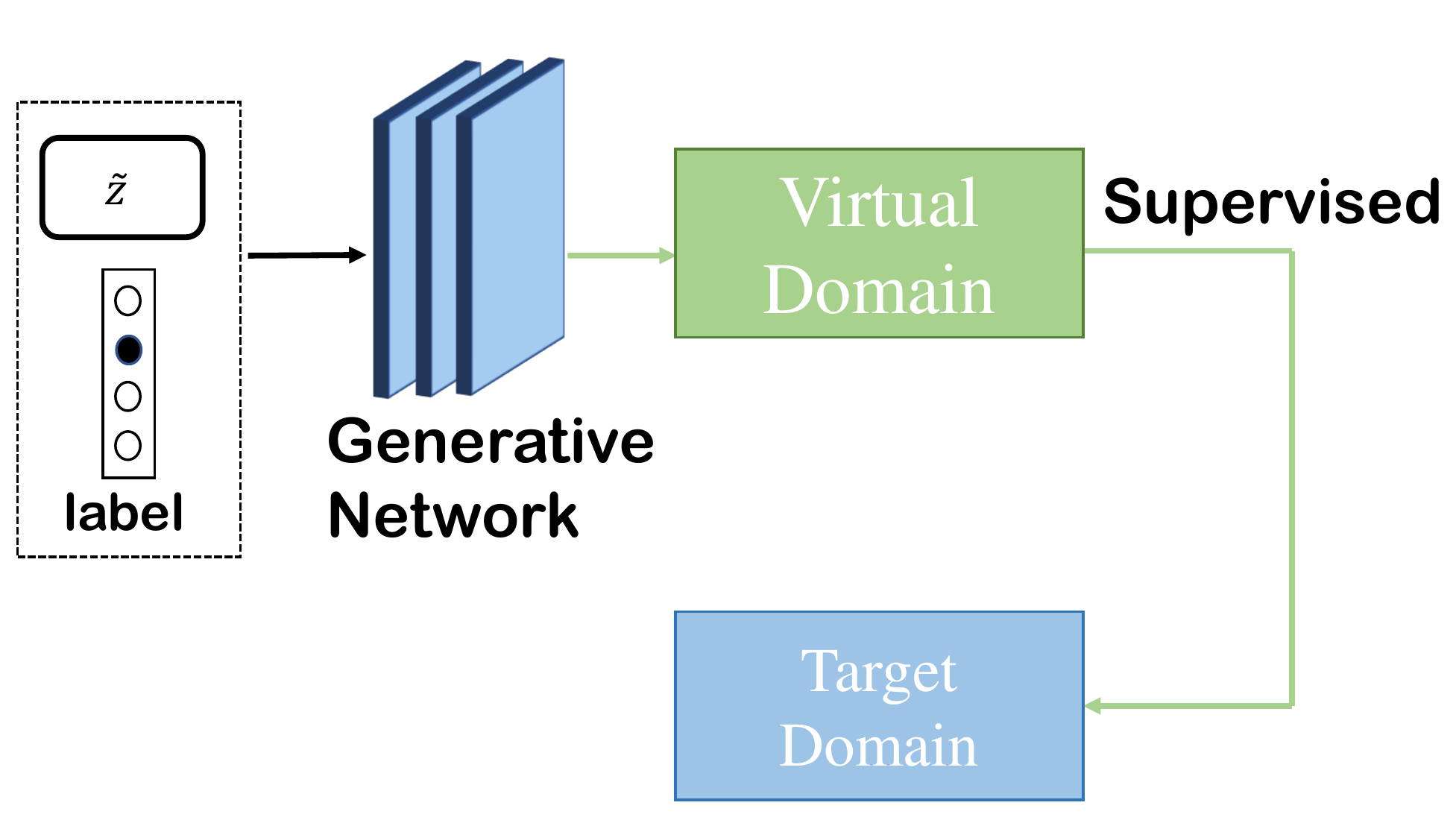}
              }
              \subfigure[Intra-domain Adversarial Alignment]{
              \includegraphics[width=0.33\linewidth] {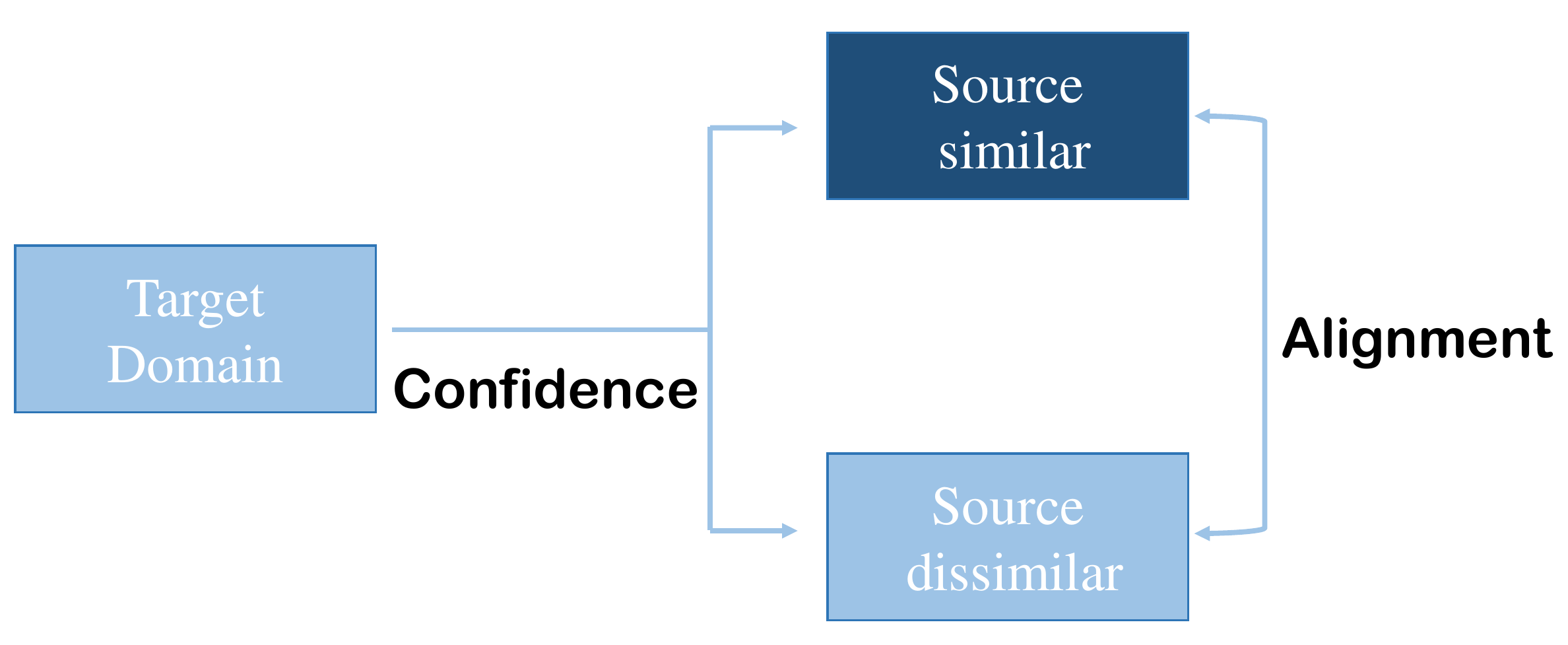}
              \label{fig:ablation}
              }
              \subfigure[Perturbed Domain Supervision]{\label{fig:4c}
              \includegraphics[width=0.3\linewidth] {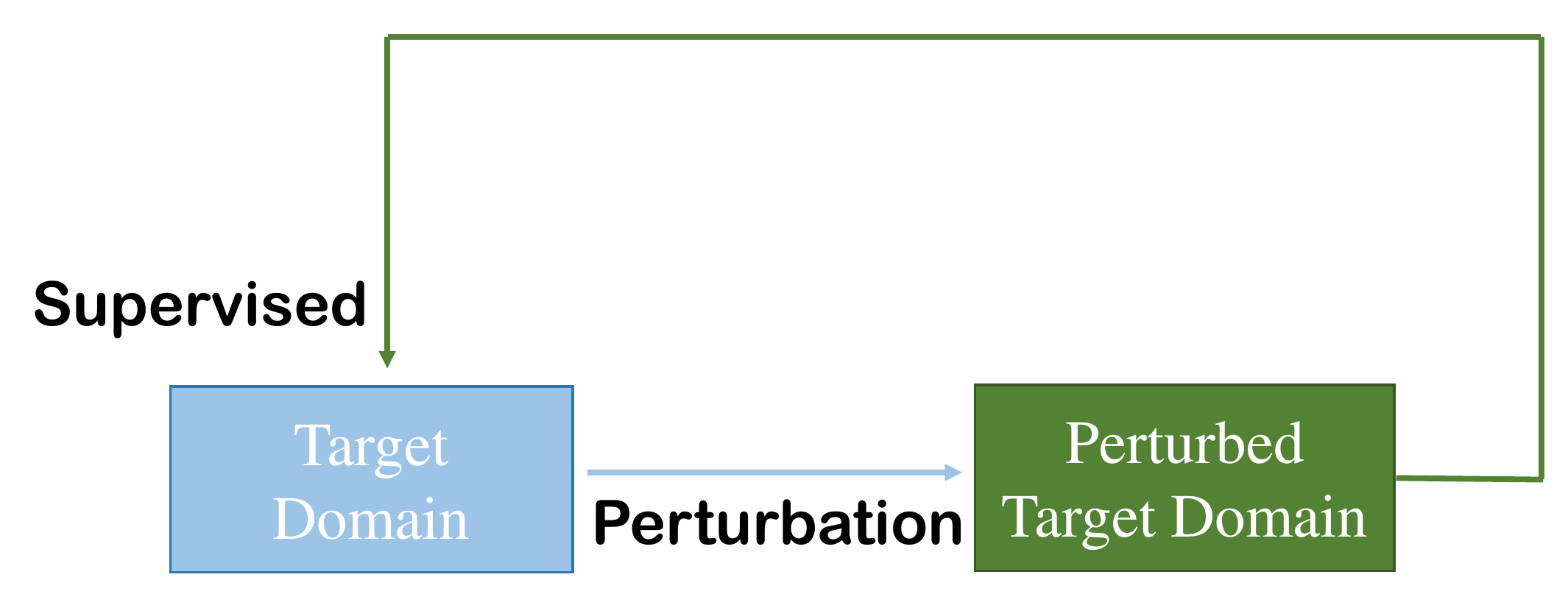}
              \label{fig:ablation}
              }
                \vspace{-7pt}
              \caption{An overview of Domain Reconstruction methods.}
              \vspace{-10pt}
              \label{fig:Domain Reconstruction}
          \end{figure*}

\section{Data-based methods}
\label{sec:data_methods}


The research of this broad class of data-based approaches can be divided into two routes based two different motivations. On the one hand, consider the unavailability of the source domain data, one of the most intuitive ideas is to mimic the source domain data through the source domain information implied in the model, or to reconstruct an intermediate domain or another domain to compensate for the absence of the source domain data. On the other hand, consider the unannotation of the target domain, another class of methods can be seen as trying to explore the potential data structure or clustering information in the unlabeled target domain data, making it possible to perform the domain adaptation task independently with the target domain data alone. In our categorization, we call the above two directions as Domain-based Reconstruction and Image-based Information Extraction, respectively. In the following, we describe each of them separately in detail.

\subsection{Domain-based Reconstruction}
The core purpose of this class of methods is to reconstruct a new domain to supervise the target domain. In general, there are three directions to construct a new domain, one is to generate a virtual source domain, one is to construct a perturbed target domain for robust learning, and the last one is to divide the target domain into source-like and target-like parts and then perform an intra-domain adversarial to achieve alignment.

\subsubsection{Virtual Domain Generation}

In unsupervised domain adaptation, a typical paradigm is to supervise the model on the source domain, and meanwhile employing some techniques to align the distributions of the source and target domains directly. The overall training goal of this process can be formulated as:

\begin{equation}
       \begin{aligned}
    \mathcal{L}_{UDA_M}=\mathcal{L}_{cls}\left(\mathcal{C}\left(\mathcal{F}\left(\mathcal{X}^{s}\right)\right), \mathcal{Y}^{s}\right) + \mathcal{L}_{div}\left(\mathcal{F}\left(\mathcal{X}^{s}\right),\mathcal{F}\left(\mathcal{X}^{t}\right)\right),
       \end{aligned}
   \end{equation}
where $\mathcal{L}_{cls}$ is the supervised loss in the source domain, usually using cross-entropy. $\mathcal{L}_{div}$ denotes the criterion used for alignment. However, in SFDA, $\mathcal{X}^{s}$ is not available during adaptation. Therefore, the virtual domain generation methods attempt to construct a domain to compensate for the absence of source data. The above equation can be extended as follows:

\begin{equation}
    \mathcal{L}_{SFDA}=\mathcal{L}_{gen}\left(\mathcal{C}\left(\mathcal{F}\left(\mathcal{X}^{v}\right)\right)\right) + \mathcal{L}_{div}\left(\mathcal{F}\left(\mathcal{X}^{v}\right),\mathcal{F}\left(\mathcal{X}^{t}\right)\right),
 \end{equation}
 where $\mathcal{X}^{v}$ denotes the virtual domain and  $\mathcal{L}_{gen}$ denotes the generative loss of the virtual domain. Note that the model has been pre-trained on the source domain, so it is possible to simulate the domain distribution by exploring the outputs of the model.
 
 It can be found that compared to Eq.~\ref{eq:1}, SFDA should be concerned with the quality of virtual domain generation in addition to the alignment between domains. Since when the virtual domain is generated, the alignment problem of SDFA becomes similar to that of UDA, e.g., ADDA \cite{tzeng2017adversarial} is used in VDA-DA \cite{tian2021vdm} and MMD~\cite{gretton2006kernel} is employed as an alignment metric in STDA~\cite{tian2021source}, we concentrate on how to generate the virtual domain in this section. Technically, the virtual domains can be generated in two ways: based on adversarial generation or based on Gaussian distribution.
 
 Drawing on the idea of GAN~\cite{goodfellow2020generative}, the adversarial generation can be expressed as:
 
 \begin{equation} \label{eq:1}
       \begin{aligned}
    \mathcal{L}\left(\mathcal{C},\mathcal{F},\mathcal{D},G\right)=\mathcal{L}_{adv} + \mathcal{L}_{con},
       \end{aligned}
   \end{equation}
 where $\mathcal{L}_{con}$ is the consistency loss used to constrain the semantic consistency across domains and can contain multiple perspectives. $\mathcal{L}_{adv}$ represents the adversarial loss between the generator and the domain discriminator, and is usually based on the following equation:
 
 \begin{equation}
\begin{aligned}
&\mathcal{L}_{adv}\left(G\right) = \mathbb{E}_{y,\widetilde{z}}\left[\log \mathcal{D} \left(1-G\left(y,\widetilde{z}\right)\right)\right], \\
\mathcal{L}_{adv}\left(\mathcal{D}\right) &= \mathbb{E}_{x_{t}\sim \mathcal{X}_{t}}\left[\log\mathcal{D}\left(x_{t}\right)\right]+\mathbb{E}_{y,\widetilde{z}}\left[\log \left(1- \mathcal{D}\left(G\left(y,\widetilde{z}\right)\right)\right)\right],
  \end{aligned}
\end{equation}
where $\widetilde{z}$ stands for the noise vector, $G\left(y,\widetilde{z}\right)$ represents the generator conditioned on a pre-defined labeled $y$, which differs from the traditional generator \cite{hoffman2018cycada,liu2016coupled}.

The focus of this class of methods lies in the design of $\mathcal{L}_{con}$. 3C-GAN \cite{li2020model} is a pioneering work in this area, for the generator, it emphasizes the semantic similarity between the generated $x_v=G\left(y,\widetilde{z}\right)$ and the label $y$. For the classifier, a deterministic constraint, a weight constraint, and a clustering-based constraint have been proposed to improve the performance. Further, SDDA \cite{kurmi2021domain} proposes a consistency loss based on the domain discriminator, which guides the feature extractor to extract domain-invariant features by a binary classification loss. CPGA \cite{qiu2021source} is inspired by InfoNCE \cite{oord2018representation} and generates more representative prototypes for each category by adding a contrastive loss.

Another way of virtual domain generation is to simulate the distribution of the source domain based on Gaussian distribution according to the implicit knowledge contained in the pre-trained model. Generall, there are mainly two views. One view is to sample noises from standard Gaussian distribution as the inputs of the generator, then the virtual source data generation process $\widetilde{x}$ can be expressed as:

\begin{equation}
       \begin{aligned}
    \widetilde{x} = G\left(\widetilde{z}\right), \widetilde{z}\sim \mathcal{N}\left(0,1\right).
       \end{aligned}
   \end{equation}

Based on data-free knowledge distillation \cite{yin2020dreaming}, Liu \emph{et al.} \cite{liu2021source} propose a batch normalization statistical loss to model the source domain distribution using the mean and variance stored in the BN layer of the source domain model. 

An alternative view is to regard the source domain distribution as a mixture of multiple Gaussian distributions. For instance, VDM-DA \cite{tian2021vdm} constructs a Gaussian mixture model to represent the distribution of the virtual domain in the feature space as:

\begin{equation}
       \begin{aligned}
   D^{v}\left(f_{v}\right)=\sum_{k=1}^{K}\pi_{k}\mathcal{N}\left(f_{v}\vert \mu_{k}, \sigma^{2}I\right),
       \end{aligned}
   \end{equation}
where $f_v$ stands for the virtual features and $\pi_{k}$ denotes the mixing coefficients with $\sum_{k=1}^{K}\pi_{k}=1$. To estimate the parameters of the Gaussian mixture model, VDM-DA empirically sets $K$ to the number of categories and considers that the weights of the source classifier implicitly contain prototypical information about each category and derives the mean and standard deviation of the model based on the source classifier. SoFA \cite{yeh2021sofa} models the inference process and the generation process, respectively, and derives the mixture of Gaussian distribution from the predicted classes as the reference distribution.

However, using generative models for virtual domain is not only costly but also difficult to perform the domain generalization well particularly when the underlying data pattern is complex. Therefore, some approaches attempt to use a non-generative approach, e.g., to construct a virtual source domain by selecting some reliable data directly from the target domain. One most common idea is that by feeding the target domain images into the source model, samples with high prediction entropy are considered closer to the source domain distribution and can be used to represent the source domain distribution to some extent. Another key issue raised by this method, however, is the problem of insufficient data for the virtual source domain. For example, Du \emph{et al.} \cite{du2021generation} found that the number of samples that can be selected to construct the virtual source domain is only one-tenth of the target domain, which is not sufficient to support a data distribution that represents the entire source domain.

To tackle this problem, a range of approaches focus on how to scale up virtual source domain data. Mixup \cite{zhang2017mixup} can be regarded as an effective technique and PS \cite{du2021generation} mixes the samples in the virtual source domain with each other, which can be expressed as:

\begin{equation}
       \begin{aligned}
        \widetilde{x}_{s,aug} = \lambda \widetilde{x}_{s}^{i} + \left(1-\lambda\right)\widetilde{x}_{s}^{j},\\
        \widetilde{y}_{s,aug} = \lambda \widetilde{y}_{s}^{i} + \left(1-\lambda\right)\widetilde{y}_{s}^{j},\\
       \end{aligned}
   \end{equation}
where $\lambda$ denotes the mixup coefficient, $\widetilde{x}_{s}^{i}$,$\widetilde{x}_{s}^{j}$ are the samples selected by prediction entropy from the virtual source domain and $\widetilde{y}_{s}^{i}$,$\widetilde{y}_{s}^{j}$ are their corresponding labels respectively. In this way, an augmented virtual source domain are obtained by $\widetilde{D}^{s}_{aug}=\widetilde{D}^{s}\cup \widetilde{D}^{aug}$ with $\widetilde{D}^{aug}=\{\widetilde{x}_{s,aug},\widetilde{y}_{s,aug}\}$. UIDM \cite{ma2021uncertainty}, instead, first performs a mixup in the target domain, and then takes into account not only the prediction entropy in the sample selection of the virtual source domain, but also the data uncertainty brought by the mixup operation and the model uncertainty brought by the dropout operation in a more comprehensive manner.

In addition to data augmentation by mixup, another idea is to keep transferring samples from the target domain to the virtual source domain during training. Ye \emph{et al.} \cite{ye2021source} proposes a nonlinear weight entropy minimization loss to continuously reduce the entropy value of high confidence samples while leaving low confidence samples unaffected during the training process, resulting in more reliable samples. Although the number of samples in the virtual source domain is increasing, for some difficult categories their virtual source domain samples may still be severely lacking. ProxyMix \cite{ding2022proxymix} poses this issue and constructs a class-balanced proxy source domain using the nearest neighbors of each prototype as well as a intra-domain mixup.

\subsubsection{Intra-domain Adversarial Exploration}
In the previous section, we introduced the way to handle the problem of unavailable source data by generating the virtual domain, which views SFDA as a cross-domain problem. Instead, some other methods perform a binary split within the target domain, and then conduct adversarial learning between the two data populations while maintaining the knowledge of the source model. Consequently, it achieves the consistency of the overall data distribution in the target domain, which views SFDA as an intra-domain alignment. It is worth noting that the previously introduced virtual source domain methods share certain inspiration with this class of methods in terms of confidence, but the former emphasizes on how to construct the virtual source domain, while the latter focuses on intra-domain distribution alignment in an adversarial manner. Therefore, we categorize the former into the virtual domain reconstruction class.

Technically, intra-domain adversarial alignment mostly uses diverse classifiers and the feature extractor for adversarial purpose, which can be seen as a variant of bi-classifier paradigm  adversarial methods~\cite{saito2018maximum,lee2019sliced,yu2023classification} in conventional UDA. Here we first review the generic three-step procedures of the bi-classifier paradigm:

\begin{itemize}
\item[$\bullet$] \textbf{Step 1} Source training:
\begin{equation} \nonumber
       \begin{aligned}
   \min_{\theta_{\mathcal{F}}, \theta_{\mathcal{C}_{1}}, \theta_{\mathcal{C}_2}} \sum_{i=1}^{2}\mathcal{L}_{cls}\left(\mathcal{C}_{i}\left(\mathcal{F}\left(\mathcal{X}^{s}\right)\right), \mathcal{Y}^{s}\right) ,
       \end{aligned}
   \end{equation}
where $\mathcal{L}_{cls}$ denotes the cross entropy loss, $\mathcal{C}_{1}$ and $\mathcal{C}_{2}$ re[resent the two classifiers, $\theta_{\mathcal{F}}, \theta_{\mathcal{C}_{1}}, \theta_{\mathcal{C}_2}$ is the model parameters corresponding to the feature extractor and classifiers respectively.

\item[$\bullet$] \textbf{Step 2} Maximum classifier prediction discrepancy:
\begin{equation} \nonumber
       \begin{aligned}
   \min _{\theta_{\mathcal{C}_{1}}, \theta_{\mathcal{C}_{2}}} \mathcal{L}_{cls}\left(\mathcal{X}^{s}, \mathcal{Y}^{s}\right) - \mathcal{L}_{dis}\left(\mathcal{Y}^{t}_{1}|\mathcal{X}^{t},
   \mathcal{Y}^{t}_{2}|\mathcal{X}^{t}\right),
       \end{aligned}
   \end{equation}
 where the first term is identical to step 1, $\mathcal{L}_{dis}$ refers to the discrepancy loss of the two classifiers' prediction.
\item[$\bullet$] \textbf{Step 3} Minimize classifier prediction discrepancy:
\begin{equation} \nonumber
       \begin{aligned}
   \min _{\theta_{\mathcal{F}}}          \mathcal{L}_{dis}\left(\mathcal{Y}^{t}_{1}|\mathcal{X}^{t},
   \mathcal{Y}^{t}_{2}|\mathcal{X}^{t}\right).
       \end{aligned}
   \end{equation}
\end{itemize}

We now revisit this three-step procedures in the SFDA setting. For the first step, the source model is usually trained in the same way. For the second and third steps, the $\mathcal{L}_{cls}$ term is intractable because of the missing source data, and source knowledge can not be well maintained. As a result, the prediction discrepancy term $\mathcal{L}_{dis}$ in this case becomes unanchored.

To solve the above problem in the SFDA setting, a source-specific classifier can be preserved by freezing the parameters of one classifier, and a target-specific classifier can be trained on the unlabeled target data. In addition, the target domain can be divided into source-similar and source-dissimilar samples by the source-specific classifier's prediction confidence or some other criteria, thus the intra-domain adversarial alignment can be achieved by the prediction between the target-specific classifier and the source-specific classifier on the two data populations. Formally, the intra-domain adversarial alignment methods can be generalized to two steps as:

\textbf{Step $\rm{1}^{\star}$} Update the target-specific classifier by maximizing the prediction discrepancy between the target-specific classifier and the source-specific classifier on the source-dissimilar samples while minimizing the same prediction discrepancy on the source-similar samples:

\begin{equation} \label{eq:8}
       \begin{aligned}
   \min_{\theta_{\mathcal{C}_{t}}}  \sum_{x\in \mathcal{X}^{t}_{h}}\mathcal{L}_{dis}\left(p^{s}\left(x\right),p^{t}\left(x\right)\right)-\sum_{x\in \mathcal{X}^{t}_{l}}\mathcal{L}_{dis}\left(p^{s}\left(x\right),p^{t}\left(x\right)\right),
       \end{aligned}
   \end{equation}
where $\mathcal{X}^{t}_{h}$ and $\mathcal{X}^{t}_{l}$ represent the high-confidence (source-similar) and low-confidence (source-dissimilar) samples in the target domain with $\mathcal{X}^{t}_{h}\cup\mathcal{X}^{t}_{l}=\mathcal{X}^{t}$, $p^{s}\left(x\right)$ and $p^{t}\left(x\right)$ denote the predictions of source-specific classifier and target-specific classifier, respectively.

\textbf{Step $\rm{2}^{\star}$} Update the feature extractor to pull the target features within the two classifiers' decision boundaries:

\begin{equation}
   \min_{\theta_{\mathcal{F}}}  \sum_{x\in \mathcal{X}^{t}}\mathcal{L}_{dec}\left(p^{s}\left(x\right),p^{t}\left(x\right)\right),
   \end{equation}
where $\mathcal{L}_{dec}$ stands for the decision loss to improve the generalization ability of the feature extractor. 

In general, there are a total of three directions to improve the effect of intra-domain adversarial alignment, which are the selection of $\mathcal{L}_{dis}$ and $\mathcal{L}_{dec}$, the division between $\mathcal{X}^{t}_{h}$ and $\mathcal{X}^{t}_{l}$, and the optimization of network structure. Since we summarize the general framework for in-domain adversarial alignment, we now discuss the relevant methods under this framework. 

$\rm{A^{2}Net}$~\cite{xia2021adaptive}, as the first work of intra-domain adversarial alignment, is based on a voting strategy to divide the target domain by concatenating the predictions of the source-specific and target-specific classifiers together and performing a Softmax operation. Concretely, if the voting score of the source-specific classifier is higher than the voting score of the target-specific classifier, it will be seen as like the source-similar features and vice versa. Notably it proposes a Soft-Adversarial mechanism to train the target-specific classifier and feature extractor with the exchange of voting scores, which can be expressed as:

\begin{equation} \label{eq:10}
       \begin{aligned}
   \min_{\theta_{\mathcal{C}_{t}}}  -\sum_{i=1}^{n_t}\left(\alpha_{i}^{s}\log\left(\sum_{k=1}^{K}p^{st}_{(i)k}\right)+\alpha_{i}^{t}\log\left(\sum_{k=K+1}^{2K}p^{st}_{(i)k}\right)\right),\\
   \min_{\theta_{\mathcal{F}}}  -\sum_{i=1}^{n_t}\left(\alpha_{i}^{t}\log\left(\sum_{k=1}^{K}p^{st}_{(i)k}\right)+\alpha_{i}^{s}\log\left(\sum_{k=K+1}^{2K}p^{st}_{(i)k}\right)\right),\\
       \end{aligned}
   \end{equation}
 where $\alpha_{i}^{s}$, $\alpha_{i}^{t}$ represent the voting scores of the source-specific classifier and  target-specific classifier respectively, $p^{st}=\sigma\left({[p^{s},p^{t}]}^{T}\right) \in \mathbb{R}^{2K}$ is the concentration of $p^{s}$ and $p^{t}$ after activation. Although the presentation of Eq.~\ref{eq:10} is different from that of Eq.~\ref{eq:8}, it is still essentially an adversarial training between a feature extractor and a target classifier, so our framework is still applicable.

 BAIT\cite{yang2020casting} is another typical approach that conforms to the in-domain adversarial alignment paradigm, which uses KL divergence as the discrepancy loss as well as a bite loss that raises the output entropy of the two classifiers, denoted as follows:

 \begin{equation}
     \begin{aligned}
      \min_{\theta_{\mathcal{C}_{t}}}  \sum_{x\in \mathcal{X}^{t}_{h}}&\mathcal{D}_{SKL}\left(p^{s}\left(x\right),p^{t}\left(x\right)\right)-\sum_{x\in \mathcal{X}^{t}_{l}}\mathcal{D}_{SKL}\left(p^{s}\left(x\right),p^{t}\left(x\right)\right), \\
   &\min_{\theta_{\mathcal{F}}} \sum_{i=1}^{n_t}\sum_{k=1}^{K}\left[ -p_{i,k}^{s}\log p_{i,k}^{t}-p_{i,k}^{t}\log p_{i,k}^{s}\right],
         \end{aligned}
 \end{equation}
where $\mathcal{D}_{SKL}(a,b)=\frac{1}{2}\left(\mathcal{D}_{KL}\left(a\vert b\right)+\mathcal{D}_{KL}\left(b\vert a\right)\right)$. It is worth mentioning that BAIT empirically found in its experiments that for the selected thresholds, it is better to divide the source-similar and source-dissimilar sets into the same size, we consider this to be an informative implications. 

The KL divergence serves as a relatively good measure of the difference in prediction distributions between two classifiers, but it ignores the determinacy of the classifier outputs, which may lead to ambiguous output problems. Both D-MCD \cite{chu2022denoised} and DAMC \cite{zong2022domain} are concerned with this problem and have adopted CDD distance \cite{li2021bi} to measure both consistency and determinacy of the outputs. In addition to this, they have made some improvements to the network structure. In D-MCD, Chu \emph{et al.} takes advantage of the model's tendency to remember simple samples in the early stages of the training process and designs a strong-weak paradigm to jointly filter the samples to avoid incorrect high-confidence samples. DAMC, on the other hand, uses a network structure with more than two classifiers to provide a tighter upper bound on the domain gap for classification and derives the optimal number of classifiers from being the same as the number of categories. However, all these extra network structures also increase the computational overhead.

So far, two major lines of research in UDA have been reflected in SFDA. In the virtual domain generation class, it is possible to use some metrics or adversarial-learning-based strategy to align the virtual and target domain distributions. In the intra-domain adversarial alignment, a variant of bi-classifier paradigm can be used. Both these methods indicate the potential research connection between SFDA and UDA.

 \subsubsection{Perturbed Domain Supervision}
 This class of methods is essentially based on the assumption that both source and target domain features originate from a domain-invariant feature space \cite{hoffman2016cross}, i.e., the source and target domains are actually formed by domain-invariant features plus some domain-biased factors. From the viewpoint of perturbation, it is feasible to use the target domain data to perturb the source data for semantic augmentation \cite{li2021transferable} in UDA, so as to guide the model to learn domain-invariant features. Then in the absence of source domain data, it is also possible to add some appropriate domain-related perturbations on the unlabeled target data during the model training. If the model can resist these domain-related perturbations, it is possible to make the model acquire domain-invariant features. As shown in Fig.~\ref{fig:4c}, the perturbed target domain acts as the guidance to the source domain, so we refer to it as the perturbed target domain supervision. Although there are still few approaches on this technique route, due to the distinct technical characteristic, we consider it as a category parallel with other two classes of approaches in Domain-based Reconstruction.

 As a representative method in this category, SOAP \cite{xiong2021source} argues that the ultimate goal of the model is to find domain-invariant features, and expresses the adjustment direction of the model as a vector and reflects it in UDA and SFDA as:


	\begin{equation}
		\left\{
		\begin{aligned}
			 &{\overrightarrow{\epsilon}\left(\mathcal{D}^{t},\mathcal{D}^{I}\right)}_{UDA} = \overrightarrow{\epsilon}\left(\mathcal{D}^{t},\mathcal{D}^{s}\right)+\overrightarrow{\epsilon}\left(\mathcal{D}^{s},\mathcal{D}^{I}\right),\\
    \\
			 &{\overrightarrow{\epsilon}\left(\mathcal{D}^{t},\mathcal{D}^{I}\right)}_{SFDA} = \overrightarrow{\epsilon}\left(\mathcal{D}^{t+},\mathcal{D}^{t}\right),
		\end{aligned}
		\right.
	\end{equation}
 where $\mathcal{D}^{I}$ denotes the domain-invariant feature space, and $\mathcal{D}^{t+}$ is a super target domain constructed by adding a target-specific perturbation $\widetilde{N}_{T}$ to the target domain samples as $X_{T+}^{i}=\beta X_{T}^{i} + (1-\beta)\widetilde{N}_{T}$. 
 
 SMT \cite{zhang2021source} is also based on this strategy, which dynamically updates the perturbed target domain. Although the motivation of the super target domain is attractive, the construction is still too simple, e.g., taking the mean value of the target images, which may not enough to reflect the property of the target domain.

 For the same purpose, FAUST \cite{lee2022feature} interprets this data perturbation from the perspective of uncertainty \cite{kendall2017uncertainties}, which first applies a random augmentation to the target images, and then encourages the feature extractor to extract consistent features over the target images before and after the perturbation by both epistemic uncertainty and aleatoric uncertainty. Unlike perturbing target data, VMP \cite{jing2022variational} introduces perturbations to the parameters of the model by variational Bayesian inference to maintain the model's discriminative power while performing model adaptation. AAA \cite{li2021divergence}, on the other hand, introduces adversarial perturbations \cite{szegedy2013intriguing} into domain adaptation by designing adversarial examples to attack the model, and the generation process of the adversarial examples can be represented as follows:
 \begin{equation}
     \begin{aligned}
     & \widetilde{x}_{t} = x_t + \frac{G\left(x_t\right)}{\vert\vert G\left(x_t\right) \vert\vert}_{2}\epsilon,
\\
\max_{\theta_{G}}&\frac{1}{n_t}\sum_{i=1}^{n_t} \mathcal{L}_{cls}\left(\mathcal{C}\left(\mathcal{F}\left(\widetilde{x}_{t}\right)\right),\hat{y}^{i}_{t}\right),
\end{aligned}
 \end{equation}
where $G$ denotes the generator, $\vert\vert \cdot \vert\vert$ denotes the Euclid norm, $\epsilon > 0$ is the perturbation magnitude to control that the perturbation will not destroy the samples' original semantic information and $\hat{y}^{i}_{t}$ is the pseudo label corresponding to $\widetilde{x}_{t}$. The authors argue that if the model defends against these attacks, the generalization ability of the model can be significantly improved in the process. This defense process can be formulated as:

 \begin{equation}
     \begin{aligned}
     \max_{\theta_{\mathcal{F}},\theta_{\mathcal{C}}}&\frac{1}{n_t}\sum_{i=1}^{n_t} \mathcal{L}_{cls}\left(\mathcal{C}\left(\mathcal{F}\left(\widetilde{x}_{t}\right)\right),\hat{y}^{i}_{t}\right).
         \end{aligned}
 \end{equation}
 
In general, the key point of this class of methods is the design of perturbations. The first point is that the magnitude of the perturbation should be appropriate, which means enhancing the model's generalization ability without destroying the original properties of the data or the model. The second point is that the perturbation should preferably reflect the characteristics of the target domain to accomplish the adaptation task on the target domain.

\subsection{Image-based Information Extraction}
The information contained in one image can be mainly divided into two parts \cite{gatys2016image}: one is the content information related to the image label, and the other is the style information that is often domain-specific. Neighborhood Clustering starts from the content information of the target images and builds on the observation that the target data itself has a clear structure and clustering, thus the consistency between neighboring nodes is required to increase the intra-class distance and decrease the inter-class distance in the feature space, so as to reduce the classification difficulty, since the misclassification mostly appears at the decision boundary of the model \cite{chen2019transferability}; Image Style Translation starts from the style information of the image, and converts the target image to the source-style without changing the content information, so that it can be better recognized by the source classifier.

\subsubsection{Neighborhood Clustering}

The key idea behind Neighborhood Clustering is that although the feature distribution in the target domain can be deviated from that in the source domain, they can still form clear clusters in the feature space and the features in the same clusters will come from the same category. Therefore, if the consistency between neighbor nodes in the feature space is encouraged, it enables feature points from the same clusters to move jointly towards a common category. 

G-SFDA \cite{yang2021generalized} is a pioneer job in this category, which proposes to use Local Structure Clustering for consistency constraints. And the methodology can be represented as follows:

\begin{equation} \label{eq:15}
       \begin{aligned}
  &\mathcal{L}_{LSC}=-\frac{1}{n}\sum_{i=1}^{n}\sum_{k=1}^{K}\log\left[p\left(x_i\right)\cdot \mathcal{B}_{S}\left(\mathcal{N}_{k}\right)\right]+\sum_{c=1}^{C}\rm{KL}\left(\overline{p}_c \vert\vert q_c\right),\\
  &\mathcal{N}_{\{1,\cdots,K\}}=\{\mathcal{B}_{F}^{j}\vert top-K\left(cos\left(\mathcal{F}\left(x_i\right),\mathcal{B}_{F}^{j}\right),\forall\mathcal{B}_{F}^{j}\in\mathcal{B}_{F}\right)\}, \\
  &\qquad\enspace\enspace\enspace\enspace\enspace\enspace\overline{p}=\frac{1}{n}\sum_{i=1}^{n}p_c\left(x_i\right),\rm{and} \:q_{\{c=1,\cdots,C\}}=\frac{1}{C}.
       \end{aligned}
   \end{equation}
From Eq.~\ref{eq:15}, LSC first finds the top-K features that are most similar to $x_i$ in the feature bank $\mathcal{B}_{F}$ by calculating the cosine similarity, and then constrains the prediction consistency of the prediction scores of these K features in the scores bank $\mathcal{B}_{S}$ with $x_i$. The second term of $\mathcal{L}_{LSC}$ is used to prevent the degenerated solution \cite{ghasedi2017deep,shi2012information} to encourage prediction balance. CPGA \cite{qiu2021source} simplifies this step by constraining only the normalized similarity between neighbor nodes as an auxiliary loss, which also helps improve accuracy.

Based on this, NRC \cite{yang2021exploiting} further refines the neighbor nodes into reciprocal and non-reciprocal neighbors based on whether they are the nearest neighbors to each other,i.e., $(j\in\mathcal{N}_{K}^{i}) \cap (i\in\mathcal{N}_{K}^{j})$, and builds an expanded neighborhood to aggregate more information in the common structure, thereby weighting the affinity of different neighbors as:

\begin{equation} 
       \begin{aligned}
  \mathcal{L}_{\mathcal{N}}=-\frac{1}{n_t}\sum_{i}\sum_{k\in\mathcal{N}_{K}^{i}} A_{ik}\mathcal{B}_{S,k}^{T}p_i, \\
  \mathcal{L}_{E}=-\frac{1}{n_t}\sum_{i}\sum_{k\in\mathcal{N}_{K}^{i}} \sum_{m\in E_{M}^{k}} r\mathcal{B}_{S,m}^{T}p_i, \\
       \end{aligned}
   \end{equation}
where $A_{ik}=1$ while $i$ and $k$ are reciprocal nodes, otherwise $A_{ik}= r = 0.1$. Similarly, Tang \emph{et al.} \cite{tang2022semantic} construct semantic neighbors on the manifold to portray more complete geometric information, Tian \emph{et al.} \cite{tian2022robust} combine with pseudo labeling techniques to obtain structure-preserved pseudo-label by the weighted average predictions of neighboring nodes. However, previous methods have only considered maintaining the consistency of the same clusters (reducing the intra-class distance), but ignored the dissimilarity of different clusters (increasing the inter-class distance), AaD~\cite{yang2022attracting} achieves this by defining two likelihood function:

\begin{equation} 
       \begin{aligned}
P\left(\mathbb{C}_{i}\right) = \prod_{j \in \mathbb{C}_{i}} p_{ij}=\prod_{j \in \mathbb{C}_{i}} \frac{e^{p_{i}^{T}p_j}}{\sum_{k=1}^{N_t}e^{p_{i}^{T}p_k}},\\
P\left(\mathbb{B}_{i}\right) = \prod_{j \in \mathbb{B}_{i}} p_{ij}=\prod_{j \in \mathbb{B}_{i}} \frac{e^{p_{i}^{T}p_j}}{\sum_{k=1}^{N_t}e^{p_{i}^{T}p_k}},\\
       \end{aligned}
   \end{equation}
where neighbor set $\mathbb{C}_{i}$ includes $K$-nearest neighbors of node $i$, and background set $\mathbb{B}_{i}$ includes the nodes which are not the neighbor of node $i$.  Once these two likelihood functions are obtained, constraints can be placed on both intra-cluster and inter-cluster by the negative log-likelihood as $L_{i}\left(\mathbb{C}_{i},\mathbb{B}_{i}\right)=-\log\frac{P\left(\mathbb{C}_{i}\right)}{P\left(\mathbb{B}_{i}\right)}$.

In addition, Neighborhood Clustering also appears widely in a new setting called active SFDA, i.e., selectively labeling a small portion of data, which has two main benefits: (1) a small portion of difficult data can be selectively labeled by constructing a neighbor graph, e.g., ELPT \cite{li2022source} selects ungraphable data on the basis of KNN; MHPL \cite{wang2022active} considers those data  satisfying the properties of neighbor-chaotic, individual-different, and target-like as the most effective choice; (2) the  discriminability of the target domain can also be improved by label propagation\cite{iscen2019label,ma2021semi} among neighbor nodes. Overall, the key point of Neighborhood Clustering methods lies in how to mine more structural information from the unlabeled target data,, and since unsupervised clustering is inherently suitable for SFDA settings for unavailable source data, there is still a wide scope for this class of methods in the future.

\subsubsection{Image Style Translation}

Image Style Translation \cite{luan2017deep,johnson2016perceptual} aims to render the same content into different styles, which generally contains two loss functions. The first one called content loss is used to ensure consistent semantic information before and after translation. The other one called transfer loss which is based on feature statistics of multiple intermediate layers to perform style translation. However, in previous methods, most of the feature statistics are considered in the form of Gram matrix \cite{gatys2016image} or instance normalization \cite{huang2017arbitrary}, which requires access to the source domain style data, but this is not feasible in the SFDA setting. So Hou \emph{et al.} used the mean $\mu_{stored}^{n}$ and variance $\sigma_{stored}^{n}$ of the image batches for the n-th layer stored in the Batch normalization (BN) layers \cite{ioffe2015batch} as a style representative of the source domain and proposed a source-free image style translation as follows:

\begin{equation} 
       \begin{aligned}
  &\mathcal{L}_{content}= {\vert\vert \mathcal{F}^{N}\left(\widetilde{x}_t\right)-\mathcal{F}^{N}\left(x_t\right) \vert\vert}_{2}, \\
  &\mathcal{L}_{style}\!= \!\!\frac{1}{N}\! \sum_{n=1}^{N}\! {\vert\vert \mu_{current}^{n}\!-\!\mu_{stored}^{n} \vert\vert}_2\!+\! {\vert\vert \sigma_{current}^{n}\!-\! \sigma_{stored}^{n} \vert\vert}_{2},
       \end{aligned}
   \end{equation}
where the feature maps in the source classifier have $N$ layers, $\widetilde{x}_t$ denotes the source-styled image through the generator.

CPSS \cite{zhao2022source}, on the other hand, focuses on improving the robustness of the model by augmenting the samples with diverse styles based on AdaIN \cite{huang2017arbitrary} during training. A model that is insensitive to changes in image style can be obtained and thus has a strong generalization capability. The training loss can be expressed as:

\begin{equation} 
       \begin{aligned}
  &\mathcal{L}_{intra}= \sigma\left({\widetilde{F}}_{i,j}\right)\left(\frac{F_{i,j}-\mu\left(F_{i,j}\right)}{\sigma\left(F_{i,j}\right)}\right) + \mu\left({\widetilde{F}}_{i,j}\right), \\
  &\mathcal{L}_{inter}= \sigma\left({\widetilde{F}}_{k,i,j}\right)\left(\frac{F_{k,i,j}-\mu\left(F_{k,i,j}\right)}{\sigma\left(F_{k,i,j}\right)}\right) + \mu\left({\widetilde{F}}_{k,i,j}\right),
       \end{aligned}
   \end{equation}
where $\mathcal{L}_{intra}$ and $\mathcal{L}_{inter}$ represent the intra-image style swap and inter-image style swap among $k$ images. ${\widetilde{F}}_{i,j}$ and ${\widetilde{F}}_{k,i,j}$ denote the shuffled patch that provides
the style feature and one image can be divided into $n_h \times n_w$ patches as:

\begin{equation}
\nonumber
F
=\begin{bmatrix}
F_{1,1}  &   \cdots\ &F_{1,n_w}\\
 \vdots    & \ddots  & \vdots  \\
 F_{n_h,1}  & \cdots\ & F_{n_h,n_w}\\
\end{bmatrix}
\end{equation}

Similar to CPSS, SI-SFDA \cite{ye2022alleviating} simulates learning representations from corrupted images in medical image segmentation by randomly masking some patches with a black background. How to make better use of the BN layer is also a concern for this class of methods, Ishii \emph{et al.} \cite{ishii2021source} implicitly implements style translation by aligning the statistics in the BN layer with the target domain distribution. As of now, this class of methods is still relatively rare, and is mostly found in the semantic segmentation.

\section{Model-based methods}
\label{model_methods}

Unlike data-based methods that focus on the data generation or the exploring data properties, model-based methods separate the model into several sub-modules, and then adjust the parameters to some of them for domain adaptation. In this category, self-training is the most dominant paradigm and even the most popular one in the whole SFDA research. Self-training methods mostly use auxiliary modules to improve model robustness and preserve source domain knowledge in combination with other methods, and we classify them separately in order to reflect our intention of modularizing the methods for an explicit insight in this survey.

\subsection{Self-training}

Since the supervision of the source data is unavailable, self-training, also known as self-supervised learning, makes use of the model's predictions on the unlabeled target domain to refine the model in a self-supervised manner. In SFDA, most self-training methods are carried out based on the ideas of pseudo-labeling, entropy minimization, and contrastive learning.

\subsubsection{Pseudo Labeling}

Without supervision, the most intuitive idea is to label the target samples based on the predictions of the source model, and then perform self-supervised learning based on these pseudo-labels, e.g., SHOT~\cite{liang2020we}. Specifically, this process can be divided into three steps: the first step \textbf{Prototype Generation} is to generate class prototypes based on the features of some high-confidence samples or all samples, the second step \textbf{Peseudo-label Assignment} is to specify pseudo-labels by the distance or similarity between other samples and class prototypes, and the third step  \textbf{Pseudo-label Filtering} finally filters out some noisy labels to keep the purity of the labels. It is worth noting that neither the first step nor the third step in this process necessarily exists. For example, a sample can be directly pseudo-labeled by the most possible class predicted by the source classifier, but we argue that they have a tight relationship with each other. We will expand on each step in more detail below.

\textbf{Prototype Generation.} One of the simplest ways is to select class prototypes by self-entropy \cite{kim2021domain,huang2022relative}, i.e., to select samples in each class with self-entropy greater than a certain threshold.  Ding \emph{et al.} \cite{ding2022proxymix} define the weights of the source classifier as the class prototypes and found that the mean accuracy is higher than employing entropy-criterion. However, the class prototypes selected in this way may not be representative. Similar to this, some methods \cite{xie2022active,li2022source} are inspired by active learning, and  consider target samples with higher free energy to be more representative of the target domain distribution which can be  used as class prototypes.  Another popular approach is to calculate the centroid \cite{liang2020we,wang2022cross,qu2022bmd,chen2022source} of each class based on DeepCluster \cite{caron2018deep}, which can be formulated as follows:

\begin{equation} 
       \begin{aligned}
    c_k = \frac{\sum_{x_t \in \mathcal{X}_t} \delta_k\left(\mathcal{F} \circ \mathcal{G}_t \left(x_t\right)\right)\mathcal{G}_t\left(x_t\right)}{\sum_{x_t \in \mathcal{X}_t} \delta_k\left(\mathcal{F}_{t}\left(x_t\right)\right)},
       \end{aligned}
   \end{equation}
where $c_k$ denotes the centroid of $k$-th class, $\delta_k$ denotes the $k$-th element in the soft-max operation. Some recent works~\cite{ding2022proxymix,ahmed2022cleaning} argue that using only one prototype does not fully characterize the class, so multiple prototypes are generated in each class. In general, the aim of this step is to attain a class prototype $p_k$ that can represent the distribution of each category in the target domain.

\textbf{Pseudo-label Assignment.} Once the class prototype is obtained, the pseudo-label of each sample can be obtained by comparing it with different class prototypes, either in terms of similarity or distance in the feature space:

\begin{equation} 
       \begin{aligned}
    y_{\widetilde{t}} = \arg \min_{k} \mathcal{L}_{pse} \left(\mathcal{G}\left(x_t\right),p_k\right),
       \end{aligned}
   \end{equation}
    where $\mathcal{L}_{pse}$ represents the metric function of how the pseudo-label is given. Note that when no class prototype is generated, the equation degenerates to
\begin{equation} 
       \begin{aligned}
    y_{\widetilde{t}} = \arg \min_{k} \mathcal{F} \circ \mathcal{G} \left(x_t\right),
\end{aligned}
   \end{equation}
where the pseudo-label is assigned directly by the most confident category of the model outputs. Moreover, the way of pseudo-labeling is also worth discussing. In the pioneering work SHOT \cite{liang2020we}, it is time-consuming to touch all the data before assigning the pseudo-label. Based on this, BMD \cite{qu2022bmd} proposes a dynamic pseudo labeling strategy to update the pseudo label in the process of domain adaptation. Shen \emph{et al.} \cite{shen2022benefits} label only a subset of the target domain to ensure accuracy. In this labeling process, how to maintain the balance between categories is also a major concern. Qu \emph{et al.} \cite{qu2022bmd} use the idea of multiple instance learning \cite{dietterich1997solving,li2015multiple} to form a balanced global sampling strategy. You \emph{et al.} \cite{you2021domain} set category-specific thresholds to keep the number between categories as consistent as possible, Li \emph{et al.} \cite{li2021imbalanced} noticed that the model may be biased towards most categories, thus proposing an imbalanced SFDA strategy with secondary label correction. 

\begin{figure}[t]
\centering
\includegraphics[width=0.95\linewidth]{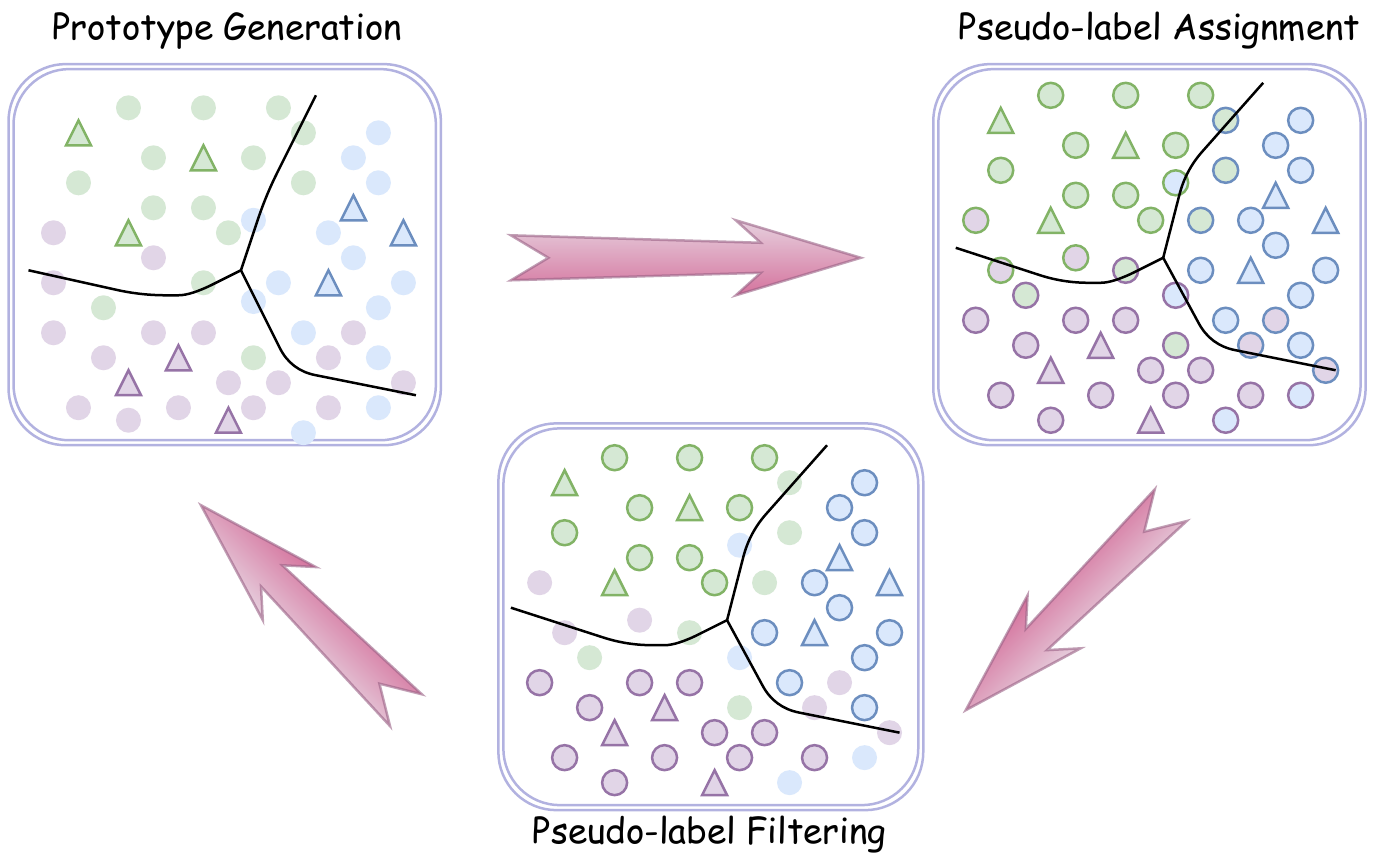}
\vspace{-10pt}
\caption{An illustration of the process in the Pseudo Labeling methods. Circles with different colors indicate samples of different classes. Circles with solid margins indicate samples with pseudo labels (mismatch between the margin color and the filled color indicates the wrong pseudo label), and triangles indicate class prototypes. The class prototypes are first generated based on the distribution of samples in the feature space, then pseudo-labels are assigned to each sample based on the class prototypes, and finally, the noisy labels are filtered. In particular, some of the methods adjust the position of the class prototypes again in the first step based on the results.}
\label{fig:pseudo_label}
\vspace{-10pt}
\end{figure}

\textbf{Pseudo-label Filtering.} Due to the presence of the domain shift, the outputs of the model inevitably contain noise, i.e., incorrect pseudo-labels, as illustrated in Fig.~\ref{fig:pseudo_label}. Therefore, the false label filtering\cite{kim2021domain,yang2022source, chu2022denoised} and the noisy label learning \cite{chen2021self,ahmed2022cleaning,liang2021source} are two important directions to improve the accuracy of pseudo labels. False label filtering is mainly to reject unreliable pseudo-labels by designing some reasonable mechanism. One is to design a specific rule mechanism, such as Kim \emph{et al.}~\cite{kim2021domain} propose an end-to-end filtering mechanism, which only recognizes a sample as reliable when its Hausdorff distance from its most similar class prototype is smaller than its distance from the second similar class prototype; another is to design an optimized network mechanism to perform filtering, such as Chu \emph{et al.} \cite{chu2022denoised} utilized the advantage that a weak model is more likely to identify hard samples \cite{arpit2017closer}, and propose an additional untrained weak network to filter on incorrect pseudo-labels. Yang \emph{et al.} \cite{yang2022source} looked at the stability of negative learning and proposed a multi-class negative learning strategy to learn the filtering threshold of pseudo-label selection adaptively. On the other hand, noise label learning is no longer just filtering wrong pseudo-labels, but also learning useful information from the noisy labels for self-refinement. SHOT++ \cite{liang2021source} is an extension of SHOT \cite{liang2020we}, attempting to improve the reliability of low confidence samples by MixMatch \cite{berthelot2019mixmatch} between high confidence samples and low confidence samples for information propagation. NEL \cite{ahmed2022cleaning} introduces ensemble learning \cite{garcia2021distillation,dong2020survey} into the SFDA setting, first performing data augmentation on the target domain samples with multiple angles based on input and feedback, and then using negative ensemble learning across multiple versions to refine the pseudo label. From the whole process of the pseudo-labeling methods, pseudo-label filtering is used as a key step to improve the accuracy of pseudo-label,  but it also brings additional resource overhead, so the efficient pseudo-label filtering technique is still worth discussing.

\subsubsection{Entropy Minimization}
Entropy minimization has profound applications in semi-supervised learning~\cite{grandvalet2004semi,jain2017subic} and unsupervised learning \cite{long2016unsupervised,springenberg2015unsupervised} methods, and was first introduced to the SFDA setting by~\cite{liang2020we}. Previous unsupervised domain adaptation mostly improves the adaptive capability of the model in the target domain by aligning the feature distributions in the source and target domains, but this is not feasible without the source domain data. Therefore, another idea is to start directly from the results, assuming that a model with adaptive ability has been obtained, then the model outputs of each sample should be deterministic, i.e., entropy minimization, and this ideal result constraint can be inversely employed to guide the optimization of the model. The entropy minimization  loss can be expressed as:

\begin{equation}  \label{eq:23}
       \begin{aligned}
    \mathcal{L}_{ent} = -\mathbb{E}_{x_t \in \mathcal{X}_t} \sum_{k=1}^{K} \delta_{k}\left(p_t\left(x_t\right)\right)\log \delta_{k}\left(p_t\left(x_t\right)\right),
\end{aligned}
   \end{equation}
where $p_t$ denotes the model's prediction with respect to sample $x_t$. Since entropy minimization loss is easy to implement and can be easily combined with other methods such as pseudo labeling methods as a loss function, image classification in the passive setting, image segmentation, medical image analysis and blind image quality assessment are rapidly adopted, it has been rapidly adopted for image classification \cite{jeon2021unsupervised,mao2021source}, image segmentation \cite{bateson2022source,kothandaraman2021ss}, medical image analysis \cite{hong2022source} and blind image quality assessment \cite{liu2022source} in the SFDA setting. However, some works \cite{krause2010discriminative,jabi2019deep} have found that entropy minimization loss may lead to a trivial solution during training, i.e., all predictions are biased towards a certain class, so the batch entropy maximization loss can also be used to ensure the diversity of predictions:

\begin{equation} \label{eq:24}
       \begin{aligned}
    \mathcal{L}_{div} = \sum_{k=1}^{K} p_k\log p_k,
\end{aligned}
   \end{equation}
where $p_k=\mathbb{E}_{x_t \in \mathcal{X}_t} \left[\delta\left(p_{t}^{k}\left(x_t\right)\right) \right]$, is to accumulate the outputs by class first in one batch, and then perform the soft-max operation.The reason for this is that within a batch, the number of samples in each category should be balanced, so that the uniform distribution has the maximum entropy value. In many SFDA methods, Eq.~\ref{eq:23} and Eq.~\ref{eq:24} are combined to form the information maximization loss:

\begin{equation} \label{eq:25}
       \begin{aligned}
    \mathcal{L}_{IM} = \mathcal{L}_{ent} + \mathcal{L}_{div}.
\end{aligned}
   \end{equation}
Based on the $\mathcal{L}_{IM}$, Ahmed  \emph{et al.} \cite{ahmed2021unsupervised} weight the different source domains in the multi-source domain setting to get the weighted information maximization, and Mao \emph{et al.} \cite{mao2021source} incorporate the neighbor node information to graph adaptation. However, except after information maximization, innovations based on entropy minimization are relatively lacking at present, and more approaches just directly use it as an auxiliary loss to improve accuracy. The key point of entropy minimization is judging from the results how to constrain the outputs of the model on the target domain from different levels such as individual and whole, deterministic and diverse, etc., and what outputs are more reasonable may be a topic worth more in-depth and detailed in the future.

\subsubsection{Contrastive Learning}

Self-supervised contrastive learning \cite{he2020momentum,chen2020simple} is a common route used for unsupervised representation learning, which focuses on learning discriminative representations by  increasing the distance between negative pairs while gathering positive pairs. This approach is characterized by the construction of positive and negative sample pairs regarding the same sample, which can be mainly categorized as the memory-bank based \cite{wu2018unsupervised}, encoder based \cite{he2020momentum} and  mini-batch based \cite{tian2020contrastive,ye2019unsupervised}. One most typical contrastive loss is the InfoNCE loss \cite{oord2018representation,he2020momentum}:

\begin{equation} 
       \begin{aligned}
    \mathcal{L}_{Info} = -\!\!\!\!\sum_{v^{+}\in V^{+}}\!\!\!\! \log \frac{\exp\left(u^{T}v^{+}/\tau\right)}{\exp \left(u^{T}v^{+}/\tau\right) + \sum_{v^{-}\in V^{-}}\exp\left(u^{T}v^{-}/\tau\right)},
\end{aligned}
   \end{equation}
where $V^{+}$,$V^{-}$ are the sets of positive and negative pairs about the same sample $u$ respectively, and $\tau$ is the temperature parameter. In unsupervised domain adaptation, the construction of positive and negative sample pairs is usually done with the target domain sample as the key value and then the source domain sample as the query. However, the latter is not available in the source-free setting, so how to construct queries that can be representative of the source domain is the central problem of this class of SFDA methods, which can be mainly divided into 3 ways. The first manner is to use the source classifier's weights as prototype features for each class of source samples \cite{wang2022cross}:

\begin{equation} 
       \begin{aligned}
    \mathcal{L}_{cdc}^{1} = -\sum_{m=1}^{M} \mathbb{I}_{\hat{y}_u=m}\log \frac{\exp\left(u^{T}w_{s}^{m}/\tau\right)}{\sum_{j=1}^{M}\exp \left(u^{T}w_{s}^{j}/\tau\right) },
\end{aligned}
   \end{equation}
where the classifier weight $W_s=[w_{s}^{1},\cdots,w_{s}^{M}]$, $\mathbb{I}_{\hat{y_t}^i=m}$ is the pseudo label indicator matrix on target sample $u$. Similarly, \cite{qiu2021source} uses the generated prototypes for contrastive learning and introduces weight loss to enhance the reliability of the pseudo label. The second way is to increase the similarity of positive pairs in the current batch for contrastive matching \cite{xia2021adaptive}:

\begin{equation} 
       \begin{aligned}
    \mathcal{L}_{cdc}^{2} = -\log \frac{\exp\left(s_{ij}\right)}{\sum_{v=1}^{b} \mathbb{I}_{v\neq i}\vert\gamma_{iv}\vert\exp \left(s_{iv}\right) },
\end{aligned}
   \end{equation}
where $s_{ij}= {[\sigma(p_{i})]}^{T}\sigma(p_{v})$ denotes the similarity of the i-th sample and the v-th sample with the soft-max operation on their model outputs $p_{i}$, $p_{v}$, $\gamma_{iv}$ is a threshold function used to determine whether the two samples belong to one class. The last class of approaches preserves a memory bank to store the outputs of the historical models, thus constructing positive and negative sample pairs \cite{huang2021model,zhao2022adaptive}. HCID \cite{huang2021model} uses the current model to encode the samples of the current batch as the query $q^{t}=M^{t}\left(x_q\right)$, and then uses the historical model to encode the previously stored samples as keys $k^{t-m}_{n}=M^{t-m}\left(x_{k_n}\right)$:

\begin{equation} 
       \begin{aligned}
    \mathcal{L}_{cdc}^{3} = -\sum_{x_q\in X_{t}}\log \frac{\exp\left(q^{T}k_{+}^{t-m}/\tau\right)\gamma_{+}^{t-m}}{\sum_{i=0}^{N} \exp \left(q^{t}k_{i}^{t-m}/\tau\right) \gamma_{i}^{t-m}},
\end{aligned}
   \end{equation}
where $N$ is the number of stored keys, $\gamma$ is the parameter used to measure the reliability of the keys, and one implementation is the classification entropy. In general, part of the processing of this class of source-free methods is similar to the basic idea of data-based SFDA, i.e., using generative or non-generative approaches to compensate for missing source data.

\subsection{Self-attention}
In computer vision tasks, localizing object regions plays an important role in domain adaptation \cite{yang2021transformer}, but traditional CNN models prefer to capture local domain-specific information, such as background information, which may not be helpful for focusing objects we really care about. Therefore, the self-attention mechanism \cite{vaswani2017attention,dosovitskiy2020image} is introduced into the SFDA setting:

\begin{equation} \label{eq:30}
       \begin{aligned}
    \mathcal{L}_{Attn_{self}}(Q,K,V)=softmax(\frac{QK^{T}}{\sqrt{d_k}})V,
\end{aligned}
   \end{equation}
where $Q$, $K$, $V$ denote query, key and value respectively, $d_k$ is the dimensions of $K$. To alleviate the context loss problem in the source domain, some methods \cite{li2022source,kothandaraman2021ss, yang2021transformer} equip the self-attentive module directly with the traditional feature extractor. Eq.~\ref{eq:30} can also be extended with cross-domain attention \cite{xu2021cdtrans}:

\begin{equation} 
       \begin{aligned}
    \mathcal{L}_{Attn_{cross}}(Q_s,K_t,V_t)=softmax(\frac{Q_{s}K^{T}_{t}}{\sqrt{d_k}})V_{t},
\end{aligned}
   \end{equation}
where $Q_s$ is from the source domain, $K_t$ and $V_t$ are from the target domain, and together they form a sample pair. Based on this, CADX \cite{bohdal2022feed} divides the target domain image into supported images and query images under the source-free setting, and improves the original patch-to-patch operation to image-to-image in order to capture the overall representations and reduce the computational burden. In addition, some approaches \cite{liu2021source,kothandaraman2023salad} further process the features from both spatial attention and channel attention perspectives to enrich the contextual semantics of the representations. Currently, attention-based SFDA methods are still relatively rare, especially transformer-based SFDA methods.

\section{Comparison}
Since the classification task constitutes the main body of existing source-free domain adaptation methods, we compare current leading SFDA methods on three most widely used classification datasets in this section. In particular, we modularize these methods according to Section~\ref{sec:data_methods} and Section~\ref{model_methods} to reflect the effectiveness of the different modules.
\subsection{Comparison Datasets}
The classification datasets used in this paper include Digits \cite{hoffman2018cycada}, VisDA-C \cite{peng2017visda}, Office-31~\cite{saenko2010adapting}, Office-Home \cite{venkateswara2017deep}, and DomainNet \cite{peng2019moment}, and we have aggregated the statistics in Table.~\ref{tab:dataset}. However, due to very little room for improvement for the Digits dataset (98.9\% in~\cite{liang2020we}) and the difficulty of the DomainNet dataset, the other three datasets are more popular in the community. Therefore, we mainly summarize the experimental results on these three datasets to represent the majority. Here are the details of these three datasets.

\textbf{Office-31} is the dominant benchmark dataset in visual transfer learning, which contains 4,652 images of 31 types of target objects commonly found in office environments, such as laptops, filing cabinets, keyboards, etc. These images are mainly derived from Amazon (online e-commerce images), Webcam (low-resolution images taken by webcam), and DSLR (high-resolution images taken by DSLR). There are 2,817 images in the Amazon dataset, with an average of 90 images per category and a single image background; 795 images in the Webcam dataset, with images exhibiting significant noise, color, and white balance artifacts; and 498 images in the DSLR dataset, with 5 objects per category and each object has taken an average of 3 times from different viewpoints.

\textbf{Office-Home} is a baseline dataset for domain adaptation that contains 4 domains, each consisting of 65 categories. The four domains are Art (artistic images in the form of drawings, paintings, decorations, etc.), Clipart (a collection of clipart images), Products (images of objects without backgrounds, and Real World (images of objects taken with a regular camera). It contains 15,500 images, with an average of about 70 images and a maximum of 99 images in each category.

\textbf{VisDA-C} is a large-scale dataset for domain adaptation from simulators to realistic environments. It consists of 12 common classes shared by three public datasets: Caltech-256 (C), ImageNet ILSVRC2012 (I), and PASCALVOC2012 (P). The source domain samples and target domain samples are synthetic images generated by rendering 3D models and real images, respectively.

\begin{table}[]
\renewcommand\arraystretch{1.3} 
\centering
\caption{Statistics of the Datasets in SFDA.}
\vspace{-8pt} \label{tab:dataset}
\renewcommand{\baselinestretch}{3}
\begin{tabular}{@{}llll@{}}
\toprule
\multicolumn{1}{c}{Dataset }                                       & Domain                 & Samples                 & Classes              \\ \midrule
\multicolumn{1}{c}{\multirow{3}{*}{Digits}} & SVHN (S)         & \multirow{3}{*}{1797}   & \multirow{3}{*}{10}  \\
\multicolumn{1}{c}{}                           & MNIST (M) &                         &                      \\
\multicolumn{1}{c}{}                           & USPS (U)      &                         &                      \\
\hline
\multicolumn{1}{c}{\multirow{3}{*}{Office}} & Amazon (A)         & \multirow{3}{*}{4652}   & \multirow{3}{*}{31}  \\
\multicolumn{1}{c}{}                           & Dslr (D) &                         &                      \\
\multicolumn{1}{c}{}                           & Webcorn (W)       &                         &                      \\
\hline
\multicolumn{1}{c}{\multirow{4}{*}{Office-Home}} & Artistic images (Ar)         & \multirow{4}{*}{15500}   & \multirow{4}{*}{65}  \\
\multicolumn{1}{c}{}                           & Clip Art (Cl) &                         &                      \\
\multicolumn{1}{c}{}                           & Product images (Pr)      &                         &                      \\
\multicolumn{1}{c}{}                           & Real-World images (Rw)      &                         &    \\   
\hline
\multicolumn{1}{c}{\multirow{2}{*}{VisDA-C}}                    & Real                   & \multirow{2}{*}{207785} & \multirow{2}{*}{12}  \\
                                               & Synthetic              &                         &                      \\
                                               \hline
\multicolumn{1}{c}{\multirow{6}{*}{DomainNet} }                    & Clipart(clp)           & \multirow{6}{*}{569010} & \multirow{6}{*}{365} \\
                                               & Infograph(inf)         &                         &                      \\
                                               & Painting(pnt)          &                         &                      \\
                                               & Quickdraw(qdr)         &                         &                      \\
                                               & Real(rel)              &                         &                      \\
                                               & Sketch(skt)            &                         &     \\
                                               \hline
\end{tabular}
\end{table}

To better represent the modules included in each method,  we categorize the SFDA methods into four classes: Domain-based Reconstruction (DR), Image-based Information Extraction (IIE), Self-Training (ST), and Self-Attention (SA). More finely, we then divide these four types of methods into nine modules, which are Virtual Domain Generation ($\mathbf{M}_{vdg}$), Intra-domain Adversarial Alignment ($\mathbf{M}_{iaa}$), Perturbed Domain Supervision ($\mathbf{M}_{pds}$), Neighborhood Clustering ($\mathbf{M}_{nc}$), Image Style Translation ($\mathbf{M}_{ist}$), Pseudo Labeling ($\mathbf{M}_{pl}$), Entropy Minimization ($\mathbf{M}_{em}$), Contrastive Learning ($\mathbf{M}_{cl}$) and Self-Attention ($\mathbf{M}_{sa})$. 

\subsection{Comparison Results}

 We summarize the classification accuracy results on the Office, Office-Home and VisDA-C datasets in Table.~\ref{tab:office}, Table~\ref{tab:office-home}, and Table~\ref{tab:visda}. It is important to note that all methods are uniformly set to the single-source homogeneous closed-set source-free domain adaptation and we cite the best results presented in the original paper. In the table, we provide the accuracy of the source-only model \cite{he2016deep} (ResNet-50 on Office, Office-Home, and ResNet-101 on VisDA-C) and the target-supervised model at the top, then we split the jobs for the years 2020, 2021, and 2022 with horizontal lines from top to bottom. Our analysis of the comparison results is provided below.

\begin{table*}[]
\renewcommand\arraystretch{1.3} 
            \centering
            \caption{Classification Accuracy (\%) Comparison for Source-free Domain Adaptation Methods on the Office-31 Dataset (ResNet-50).} \label{tab:office}
            \vspace{-8pt}
 \resizebox{\textwidth}{!}{\begin{tabular}{c|cccc|c|cccccc|c}
 \toprule
Method (Source$\rightarrow$Target) & DR & IIE & ST & SA & Modules & 
A $\rightarrow$ W & D $\rightarrow$ W & W $\rightarrow$ D & A $\rightarrow$ D & D $\rightarrow$ A & W $\rightarrow$ A & Avg. \\ \hline
Source-only \cite{he2016deep} &- & -  &    -  & - &    -  &
   68.4   & 96.7 & 99.3 & 68.9 & 62.5 & 60.7 & 76.1 \\
Target-supervised \cite{liang2021source} &- & -  &-      & - &    -  &   
 98.7   & 98.7 & 98.0 & 98.0 & 98.7 & 86.0 & 94.3 \\ \hline
SHOT \cite{liang2020we} & &   &   \Checkmark   &  &  $\mathbf{M}_{pl}+\mathbf{M}_{em}$     &
  90.1 & 98.4 & 99.9 & 94.0 & 74.7 & 74.3 & 88.6 \\
3C-GAN \cite{li2020model}  &\Checkmark &   &      &  &    $\mathbf{M}_{iaa}$     &
93.7 & 98.5 & 99.8 & 92.7 & 75.3 & 77.8 & 89.6 \\
BAIT \cite{yang2020casting} &\Checkmark &   &      &  &    $\mathbf{M}_{iaa}$     &
94.6 & 98.1 & \textbf{100.0} & 92.0 & 74.6 & 75.2 & 89.4 \\ \hline
SHOT++ \cite{liang2021source} & &   &   \Checkmark   &  &  $\mathbf{M}_{pl}+\mathbf{M}_{em}$     &
  90.4 & 98.7 & 99.9 & 94.3 & 76.2 & 75.8 & 89.2 \\
Kim \emph{et al.} \cite{kim2021domain}    & &  &   \Checkmark    &  &    $\mathbf{M}_{pl}$     &
 91.1 & 98.2 & 99.5 & 92.2 & 71.0 & 71.2 & 87.2 \\
$\mathrm{A^{2}Net}$ \cite{xia2021adaptive}    &\Checkmark &  &   \Checkmark    &  &    $\mathbf{M}_{iaa}+\mathbf{M}_{cl}$     &
 94.0 & 99.2 & \textbf{100.0} & 94.5 & 76.7 & 76.1 & 90.1 \\
 NRC  \cite{yang2021exploiting}    & &\Checkmark  &     &  &    $\mathbf{M}_{nc}$     &
 90.8 & 99.0 & \textbf{100.0} & 96.0 & 75.3 & 75.0 & 89.4 \\
  ASL \cite{yan2021augmented} & &   &   \Checkmark   &  &  $\mathbf{M}_{pl}+\mathbf{M}_{em}$     &
 94.1 & 98.4 & 99.8 & 93.4 & 76.9 & 75.0 & 89.5 \\
 TransDA \cite{yang2021transformer}    & &  &  \Checkmark   & \Checkmark &    $\mathbf{M}_{pl}+\mathbf{M}_{sa}$     &
 95.0 & \textbf{99.3} & 99.6 & \textbf{97.2} & 73.7 & \textbf{79.3} & \textbf{90.7} \\
 CPGA \cite{qiu2021source} & & \Checkmark &   \Checkmark    &  &    $\mathbf{M}_{nc}+\mathbf{M}_{pl}+\mathbf{M}_{cl}$     &
 94.1 & 98.4 & 99.8 & 94.4 & 76.0 & 76.6 & 89.9 \\
 AAA \cite{li2021divergence} & \Checkmark &   & \Checkmark     &  & 
$\mathbf{M}_{pds}+\mathbf{M}_{pl}+\mathbf{M}_{cl}$     &
94.2 & 98.1 & 99.8 & 95.6 & 75.6 & 76.0 & 89.9 \\
 VDM-DA \cite{tian2021vdm}  & \Checkmark&  & \Checkmark     &  &    $\mathbf{M}_{vdg}+\mathbf{M}_{em}$     &
 94.1 & 98.0 & \textbf{100.0} & 93.2 & 75.8 & 77.1 & 89.7 \\ 
 HCL+SHOT \cite{huang2021model} &  &  & \Checkmark     &  &    $\mathbf{M}_{pl}+\mathbf{M}_{cl}$     &
92.5 & 98.2 & \textbf{100.0} & 94.7 & 75.9 & 77.7 & 89.8 \\\hline
 AaD \cite{yang2022attracting}  & &  \Checkmark &      &  &    $\mathbf{M}_{nc}$     &
  92.1 & 99.1 & \textbf{100.0} & 96.4 & 75.0 & 76.5 & 89.9 \\
U-SFAN \cite{roy2022uncertainty} &\Checkmark &   &   \Checkmark   &  &    $\mathbf{M}_{vdg}+\mathbf{M}_{em}$     &
92.8 & 98.0 & 99.0 & 94.2 & 74.6 & 74.4 & 88.8 \\
CoWA-JMDS \cite{lee2022confidence} &\Checkmark &   &   \Checkmark   &  &    $\mathbf{M}_{vdg}+\mathbf{M}_{pl}$     &
95.2 & 98.5 & 99.8 & 94.4 & 76.2 & 77.6 & 90.3 \\
CDCL \cite{wang2022cross} & &   &   \Checkmark   &  &    $\mathbf{M}_{pl}+\mathbf{M}_{cl}$     &
92.1 & 98.5 & \textbf{100.0} & 94.4 & 76.4 & 74.1 & 89.3 \\
D-MCD \cite{chu2022denoised} & \Checkmark&   &   \Checkmark   &  &    $\mathbf{M}_{iaa}+\mathbf{M}_{pl}$     &
93.5 & 98.8 & \textbf{100.0} & 94.1 & 76.4 & 76.4 & 89.9 \\
 BMD + SHOT++ \cite{qu2022bmd} & &   &   \Checkmark   &  &  $\mathbf{M}_{pl}+\mathbf{M}_{em}$     &
 94.2 & 98.0 & \textbf{100.0} & 96.2 & 76.0 & 76.0 & 90.1 \\
ProxyMix \cite{ding2022proxymix} &\Checkmark &   &   \Checkmark   &  & 
$\mathbf{M}_{vdg}+\mathbf{M}_{pl}$     &
 \textbf{96.7} & 98.5 & 99.8 & 95.4 & 75.1 & 75.4 & 85.6 \\
 UTR \cite{pei2022uncertainty}  & &   &   \Checkmark   &  & 
 $\mathbf{M}_{pl}$     &
93.5 & 99.1 & \textbf{100.0} & 95.0 & 76.3 & 78.4 & 90.3 \\
SCLM \cite{tang2022semantic} & & \Checkmark  &   \Checkmark   &  & 
$\mathbf{M}_{nc}+\mathbf{M}_{pl}$     &
90.0 & 98.9 & \textbf{100.0} & 95.8 & 75.5 & 76.0 & 89.4 \\
Jing \emph{et al.} \cite{jing2022variational} & \Checkmark &   &      &  & 
$\mathbf{M}_{pds}$     &
93.3 & 98.6 & \textbf{100.0} & 96.2 & 75.4 & 76.9 & 90.0 \\
 Kundu \emph{et al.} + SHOT++ \cite{kundu2022balancing} & \Checkmark   &   & \Checkmark   &    &    $\mathbf{M}_{vdg} + \mathbf{M}_{pl}$      &
 93.2 & 98.9 & \textbf{100.0} & 94.6 & \textbf{78.3} & 78.9 & \textbf{90.7} \\

 \bottomrule
\end{tabular}}
\end{table*}

 \textbf{The volume of work on SFDA is increasing quickly.}There is an explosive growth trend of the number of papers on SFDA research since 2020. On the three datasets, the highest accuracies of the SFDA methods are only 3.6\% (Kundu \emph{et al.} + SHOT++ \cite{kundu2022balancing} \cite{kundu2022concurrent} on Ofiice), 2.7\% (TransDA \cite{yang2021transformer}  on Office-Home) and 0.9\% (BMD + SHOT++ \cite{qu2022bmd} on VisDA-C) lower than the accuracy under the target supervision (oracle). To some extent, SFDA shows a tendency to outperform unsupervised domain adaptation methods, both in terms of the number of works and accuracy, which not only reflects the great potential of this field, but also indicates the need for a comprehensive SFDA survey.
 
 \textbf{Self-training is still the most popular research line at the moment.} Most of the SFDA methods involve the strategy of self-training. In addition, Entropy minimization \cite{liang2020we,liang2021source,roy2022uncertainty,qu2022bmd,lee2022feature} as an auxiliary loss can be easily combined with most SFDA methods to improve the discriminability of the target domain representation. Besides, contrastive learning \cite{xia2021adaptive,qiu2021source,li2021divergence} generally works with the pseudo-labeling module to play the role of domain alignment. In the absence of source data and target labels, self-supervision by means of pseudo-labeling for the target data is the most common method. However, this inevitably introduces the problem of noisy label and error accumulation. In this regard, some approaches translate the SFDA problem into the noisy label learning problem \cite{ahmed2022cleaning,chu2022denoised} to improve the model performance. There are some approaches \cite{kundu2022concurrent,pei2022uncertainty} that try to bypass the noisy label problem and implicitly regularize the self-training direction of the target domain, achieving satisfying results. From the experimental results on two of the three datasets, the methods that achieve the highest accuracy all use the pseudo labeling method SHOT++ \cite{liang2021source} as the baseline. For instance, Kundu \emph{et al.}\cite{kundu2022balancing} achieves 90.7\% on Office, and BMD \cite{qu2022bmd} achieves 88.7\% on VisDA. These observations show that pseudo labeling is a versatile and effective technique, which is also a strong baseline for improvement.

 \begin{table*}[]
\renewcommand\arraystretch{1.3} 
            \centering
            \caption{Classification Accuracy (\%) Comparison for Source-free Domain Adaptation Methods on the Office-Home Dataset (ResNet-50).} \label{tab:office-home}
            \vspace{-8pt}
 \resizebox{\textwidth}{!}{\begin{tabular}{c|cccc|c|cccccccccccc|c}
 \toprule
Method  & DR & IIE & ST & SA & Modules & Ar $\rightarrow$ Cl & Ar $\rightarrow$ Pr & Ar $\rightarrow$ Rw & Cl $\rightarrow$ Ar & Cl $\rightarrow$ Pr & Cl $\rightarrow$ Rw & Pr $\rightarrow$ Ar & Pr $\rightarrow$ Cl & Pr $\rightarrow$ Rw & Rw $\rightarrow$ Ar & Rw $\rightarrow$ Cl & Rw $\rightarrow$ Pr & Avg. \\ \hline
Source-only \cite{he2016deep} & -&  - &  -    &-  &   -   &
   34.9   & 50.0 & 58.0 & 37.4 & 41.9 & 46.2 & 38.5 & 31.2 & 60.4 & 53.9 & 41.2 &59.9 & 46.1 \\ 
Target-supervised \cite{liang2021source} & -&  - &   -   & - &  -    &
   77.9   & 91.4 & 84.4 & 74.5 & 91.4 & 84.4 & 74.5 & 77.9 & 84.4 & 74.5 & 77.9 &91.4 & 82.0 \\\hline
 SHOT \cite{liang2020we} & &   &   \Checkmark   &  &  $\mathbf{M}_{pl}+\mathbf{M}_{em}$     &
 57.1   & 78.1 & 81.5 & 68.0 & 78.2 & 78.1 & 67.4 & 54.9 & 82.2 & 73.3 & 58.8 &84.3 & 71.8 \\
 BAIT \cite{yang2020casting} &\Checkmark &   &      &  &    $\mathbf{M}_{iaa}$     &
57.4   & 77.5 & 82.4 & 68.0 & 77.2 & 75.1 & 67.1 & 55.5 & 81.9 & 73.9 & 59.5 &84.2 & 71.6 \\ \hline
 SHOT++ \cite{liang2021source} & &   &   \Checkmark   &  &  $\mathbf{M}_{pl}+\mathbf{M}_{em}$     &
  57.9   & 79.7 & 82.5 & 68.5 & 79.6 & 79.3 & 68.5 & 57.0 & 83.0 & 73.7 & 60.7 &84.9 & 73.0 \\
$\mathrm{A^{2}Net}$ \cite{xia2021adaptive}    &\Checkmark &  &   \Checkmark    &  &    $\mathbf{M}_{iaa}+\mathbf{M}_{cl}$     &
 58.4   & 79.0 & 82.4 & 67.5 & 79.3 & 78.9 & 68.0 & 56.2 & 82.9 & 74.1 & 60.5 &85.0 & 72.8 \\
 NRC \cite{yang2021exploiting}    & &\Checkmark  &     &  &    $\mathbf{M}_{nc}$     &
 57.7   & 80.3 & 82.0 & 68.1 & 79.8 & 78.6 & 65.3 & 56.4 & 83.0 & 71.0 & 58.6 &85.6 & 72.2 \\
 TransDA \cite{yang2021transformer}    & &  &  \Checkmark   & \Checkmark &    $\mathbf{M}_{pl}+\mathbf{M}_{sa}$     &
 \textbf{67.5}   & \textbf{83.3} & \textbf{85.9} & \textbf{74.0} & \textbf{83.8} & \textbf{84.4} & \textbf{77.0} & \textbf{68.0} & \textbf{87.0} & \textbf{80.5} & \textbf{69.9} &\textbf{90.0} & \textbf{79.3} \\
 G-SFDA \cite{yang2021generalized} & &\Checkmark  &     &\Checkmark  &    $\mathbf{M}_{nc}+\mathbf{M}_{sa}$     &
 57.9   & 78.6 & 81.0 & 66.7 & 77.2 & 77.2 & 65.6 & 56.0 & 82.2 & 72.0 & 57.8 &83.4 & 71.3 \\
 CPGA \cite{qiu2021source} & & \Checkmark &   \Checkmark    &  &    $\mathbf{M}_{nc}+\mathbf{M}_{pl}+\mathbf{M}_{cl}$     &
  59.3   & 78.1 & 79.8 & 65.4 & 75.5 & 76.4 & 65.7 & 58.0 & 81.0 & 72.0 & 64.4 &83.3 & 71.6 \\
AAA \cite{li2021divergence} & \Checkmark &   & \Checkmark     &  & 
$\mathbf{M}_{pds}+\mathbf{M}_{pl}+\mathbf{M}_{cl}$     &
56.7   & 78.3 & 82.1 & 66.4 & 78.5 & 79.4 & 67.6 & 53.5 & 81.6 & 74.5 & 58.4 &84.1 & 71.8 \\
PS \cite{du2021generation}  & \Checkmark &  &      &  &    $\mathbf{M}_{vdg}$     &
     57.8   & 77.3 & 81.2 & 68.4 & 76.9 & 78.1 & 67.8 & 57.3 & 82.1 & 75.2 & 59.1 &83.4 & 72.1 \\ \hline
AaD \cite{yang2022attracting}  & &  \Checkmark &      &  &    $\mathbf{M}_{nc}$     &
   59.3   & 79.3 & 82.1 & 68.9 & 79.8 & 79.5 & 67.2 & 57.4 & 83.1 & 72.1 & 58.5 &85.4 & 72.7 \\
U-SFAN \cite{roy2022uncertainty} &\Checkmark &   &   \Checkmark   &  &    $\mathbf{M}_{vdg}+\mathbf{M}_{em}$     &
    57.8   & 77.8 & 81.6 & 67.9 & 77.3 & 79.2 & 67.2 & 54.7 & 81.2 & 73.3 & 60.3 &83.9 & 71.9 \\
CoWA-JMDS \cite{lee2022confidence} &\Checkmark &   &   \Checkmark   &  &    $\mathbf{M}_{vdg}+\mathbf{M}_{pl}$     &

56.9   & 78.4 & 81.0 & 69.1 & 80.0 & 79.9 & 67.7 & 57.2 & 82.4 & 72.8 & 60.5 &84.5 & 72.5 \\
D-MCD \cite{chu2022denoised} & \Checkmark&   &   \Checkmark   &  &    $\mathbf{M}_{iaa}+\mathbf{M}_{pl}$     &
59.4   & 78.9 & 80.2 & 67.2 & 79.3 & 78.6 & 65.3 & 55.6 & 82.2 & 73.3 & 62.8 &83.9 & 72.2 \\
Kundu \emph{et al.} \cite{kundu2022concurrent} & &   &   \Checkmark   &  &  $\mathbf{M}_{pl}$     &
61.0   & 80.4 & 82.5 & 69.1 & 79.9 & 79.5 & 69.1 & 57.8 & 82.7 & 74.5 & 65.1 &86.4 & 74.0 \\
BMD + SHOT++ \cite{qu2022bmd} & &   &   \Checkmark   &  &  $\mathbf{M}_{pl}+\mathbf{M}_{em}$     &
58.1   & 79.7 & 82.6 & 69.3 & 81.0 & 80.7 & 70.8 & 57.6 & 83.6 & 74.0 & 60.0 &85.9 & 73.6 \\
ProxyMix \cite{ding2022proxymix} &\Checkmark &   &   \Checkmark   &  & 
$\mathbf{M}_{vdg}+\mathbf{M}_{pl}$     &
59.3   & 81.0 & 81.6 & 65.8 & 79.7 & 78.1 & 67.0 & 57.5 & 82.7 & 73.1 & 61.7 &85.6 & 72.8 \\
UTR \cite{pei2022uncertainty}  & &   &   \Checkmark   &  & 
$\mathbf{M}_{pl}$     &
59.8   & 81.2 & 83.2 & 67.2 & 79.2 & 80.1 & 68.4 & 56.4 & 83.0 & 73.7 & 61.2 &85.9 & 73.2 \\
FAUST \cite{lee2022feature}   & &   &   \Checkmark   &  & 
$\mathbf{M}_{pl}+\mathbf{M}_{em}$     &
61.4   & 79.2 & 79.6 & 63.3 & 76.9 & 75.2 & 65.3 & 59.4 & 79.0 & 74.7& 64.2 &86.1 & 72.0 \\
DAMC \cite{zong2022domain} & \Checkmark &   &      &  & 
$\mathbf{M}_{iaa}$     &
57.9   & 78.5 & 81.1 & 66.7 & 77.7 & 77.9 & 66.5 & 54.3 & 81.5 & 73.2 & 58.9 &84.7 & 71.6 \\
SCLM \cite{tang2022semantic} & & \Checkmark  &   \Checkmark   &  & 
$\mathbf{M}_{nc}+\mathbf{M}_{pl}$     &
58.2   & 80.3 & 81.5 & 69.3 & 79.0 & 80.7 & 69.0 & 56.8 & 82.7 &74.7 & 60.6 &85.0 & 73.1 \\
Jing \emph{et al.} \cite{jing2022variational} & \Checkmark &   &      &  & 
$\mathbf{M}_{pds}$     &
57.9   & 77.6 & 82.5 & 68.6 & 79.4 & 80.6 & 68.4 & 55.6 & 83.1 & 75.2 & 59.6 &84.7 & 72.8 \\
 Kundu \emph{et al.} + SHOT++ \cite{kundu2022balancing} & \Checkmark   &   & \Checkmark   &    &    $\mathbf{M}_{vdg} + \mathbf{M}_{pl}$      &
61.8   & 81.2 & 83.0 & 68.5 & 80.6 & 79.4 & 67.8 & 61.5 & 85.1 & 73.7 & 64.1 &86.5 & 74.5\\
 \bottomrule
\end{tabular}}
\vspace{-10pt}
\end{table*}

\begin{table*}[]
\renewcommand\arraystretch{1.3} 
            \centering
            \caption{Classification Accuracy (\%) Comparison for Source-free Domain Adaptation Methods on the VisDA-C Dataset (ResNet-101).} \label{tab:visda}
            \vspace{-8pt}
 \resizebox{\textwidth}{!}{\begin{tabular}{c|cccc|c|cccccccccccc|c}
 \toprule
Method  & DR & IIE & ST & SA & Modules & plane & bcycl & bus & car & horse & knife & mcycl & person & plant & sktbrd & train & truck & Avg. \\ \hline

Source-only \cite{he2016deep} &- &  - &   -   & - &   -   &
   55.1   & 53.3 & 61.9 & 59.1 & 80.6 & 17.9 & 79.7 & 31.2 & 81.0 & 26.5 & 73.5 &8.5 & 52.4 \\ 
Target-supervised \cite{liang2021source} & -&  - &   -   & - &  -    &   
97.0   & 86.6 & 84.3 & 88.7 & 96.3 & 94.4 & 92.0 & 89.4 & 95.5 & 91.8 & 90.7 &68.7 & 89.6 \\ \hline
SHOT \cite{liang2020we} & &   &   \Checkmark   &  &  $\mathbf{M}_{pl}+\mathbf{M}_{em}$     &
   94.3   & 88.5 & 80.1 & 57.3 & 93.1 & 94.9 & 80.7 & 80.3 & 91.5 & 89.1 & 86.3 &58.2 & 82.9 \\
3C-GAN \cite{li2020model}  &\Checkmark &   &      &  &    $\mathbf{M}_{iaa}$     &
 95.7   & 78.0 & 69.0 & 74.2 & 94.6 & 93.0 & 88.0 & 87.2 & 92.2 & 88.8 &85.1  &54.3 & 83.3 \\
BAIT \cite{yang2020casting} &\Checkmark &   &      &  &    $\mathbf{M}_{iaa}$     &
-  & - & - & - & - & - & - & - & - & - & - & - & 83.0 \\
Hou \emph{et al.} \cite{hou2020source}    &  &  \Checkmark  & \Checkmark   &    &    $\mathbf{M}_{ist} + \mathbf{M}_{em}$      &
-  & - & - & - & - & - & - & - & - & - & - & - & 63.5 \\ \hline
 SHOT++ \cite{liang2021source} & &   &   \Checkmark   &  &  $\mathbf{M}_{pl}+\mathbf{M}_{em}$     &
  97.7   & 88.4 & \textbf{90.2} & 86.3 & \textbf{97.9} & \textbf{98.6} & \textbf{92.9} & 84.1 & \textbf{97.1} & 92.2 &93.6  &28.8 & 87.3 \\
Kim \emph{et al.} \cite{kim2021domain}    &  &   & \Checkmark   &    &    $\mathbf{M}_{pl}$      &
    86.9   & 81.7 & 84.6 & 63.9 & 93.1 & 91.4 & 86.6 & 71.9 & 84.5 & 58.2 &74.5  &42.7 & 76.7 \\
$\mathrm{A^{2}Net}$ \cite{xia2021adaptive}    &\Checkmark &  &   \Checkmark    &  &    $\mathbf{M}_{iaa}+\mathbf{M}_{cl}$     &
 94.0   & 87.8 & 85.6 & 66.8 & 93.7 & 95.1 & 85.8 & 81.2 & 91.6 & 88.2 & 86.5 &56.0 & 84.3 \\
 NRC \cite{yang2021exploiting}    & &\Checkmark  &     &  &    $\mathbf{M}_{nc}$     &
 96.8   & 91.3 & 82.4 & 62.4 & 96.2 & 95.9 & 86.1 & 80.6 & 94.8 & 94.1 & 90.4 &68.2 & 85.9 \\
 G-SFDA \cite{yang2021generalized} & &\Checkmark  &     &\Checkmark  &    $\mathbf{M}_{nc}+\mathbf{M}_{sa}$     &
 96.1   & 88.3 & 85.5 & 74.1 & 97.1 & 95.4 & 89.5 & 79.4 & 95.4 & 92.9 & 89.1 &42.6 & 85.4 \\
  ASL \cite{yan2021augmented} & &   &   \Checkmark   &  &  $\mathbf{M}_{pl}+\mathbf{M}_{em}$     &
 97.3   & 85.3 & 86.9 & 70.7 & 96.4 & 72.8 & 93.0 & 80.1 & 95.5 & 78.1 & 87.7 &50.3 & 82.8 \\
 TransDA \cite{yang2021transformer}    & &  &  \Checkmark   & \Checkmark &    $\mathbf{M}_{pl}+\mathbf{M}_{sa}$     &
 97.2   & 91.1 & 81.0 & 57.5 & 95.3 & 93.3 & 82.7 & 67.2 & 92.0 & 91.8 & 92.5 &54.7 & 83.0 \\
 CPGA \cite{qiu2021source}  & & \Checkmark &   \Checkmark    &  &    $\mathbf{M}_{nc}+\mathbf{M}_{pl}+\mathbf{M}_{cl}$     &
 95.6   & 89.0 & 75.4 & 64.9 & 91.7 & 97.5 & 89.7 & 83.8 & 93.9 & 93.4 & 87.7 &69.0 & 86.0 \\
 AAA \cite{li2021divergence} & \Checkmark &   & \Checkmark     &  & 
$\mathbf{M}_{pds}+\mathbf{M}_{pl}+\mathbf{M}_{cl}$     &
94.4   & 85.9 & 74.9 & 60.2 & 96.0 & 93.5 & 87.8 & 80.8 & 90.2 & 92.0 & 86.6 &68.3 & 84.2 \\
 VDM-DA \cite{tian2021vdm}  & \Checkmark&  & \Checkmark     &  &    $\mathbf{M}_{vdg}+\mathbf{M}_{em}$     &
 96.9   & 89.1 & 79.1 & 66.5 & 95.7 & 96.8 & 85.4 & 83.3 & 96.0 & 86.6 & 89.5 &56.3 & 85.1 \\
PS \cite{du2021generation}  & \Checkmark &  &      &  &    $\mathbf{M}_{vdg}$     &
    95.3   & 86.2 & 82.3 & 61.6 & 93.3 & 95.7 & 86.7 & 80.4 & 91.6 & 90.9 & 86.0 &59.5 & 84.1 \\ \hline
 AaD \cite{yang2022attracting}  & &  \Checkmark &      &  &    $\mathbf{M}_{nc}$     &
  97.4   & 90.5 & 80.8 & 76.2 & 97.3 & 96.1 & 89.8 & 82.9 & 95.5 & 93.0 & \textbf{92.0} &64.7 & 88.0 \\
  U-SFAN \cite{roy2022uncertainty} &\Checkmark &   &   \Checkmark   &  &    $\mathbf{M}_{vdg}+\mathbf{M}_{em}$     &
   -  & - & - & - & - & - & - & - & - & - & - & - & 82.7 \\
   HCL+SHOT \cite{huang2021model} &  &  & \Checkmark     &  &    $\mathbf{M}_{pl}+\mathbf{M}_{cl}$     &
   93.3   & 85.4 & 80.7 & 68.5 & 91.0 & 88.1 & 86.0 & 78.6 & 86.6 & 88.8 & 80.0 &74.7 & 83.5 \\
CoWA-JMDS \cite{lee2022confidence} &\Checkmark &   &   \Checkmark   &  &    $\mathbf{M}_{vdg}+\mathbf{M}_{pl}$     &
96.2   & 89.7 & 83.9 & 73.8 & 96.4 & 97.4 & 89.3 & 86.8 & 94.6 & 92.1 & 88.7 &53.8 & 86.9 \\
NEL \cite{ahmed2022cleaning} & &   &   \Checkmark   &  &    $\mathbf{M}_{pl}$     &
94.5   & 60.8 & 92.3 & 87.3 & 87.3 & 93.2 & 87.6 & \textbf{91.1} & 56.9 & 83.4 & 93.7 &\textbf{86.6} & 84.2 \\
CDCL \cite{wang2022cross} & &   &   \Checkmark   &  &    $\mathbf{M}_{pl}+\mathbf{M}_{cl}$     &
97.3   & 90.5 & 83.2 & 59.9 & 96.4 & 98.4 & 91.5 & 85.6 & 96.0 & 95.8 & 92.0 &63.8 & 87.5 \\
D-MCD \cite{chu2022denoised} & \Checkmark&   &   \Checkmark   &  &    $\mathbf{M}_{iaa}+\mathbf{M}_{pl}$     &
97.0   & 88.0 & 90.0 & 81.5 & 95.6 & 98.0 & 86.2 & 88.7 & 94.6 & 92.7 & 83.7 &53.1 & 87.5 \\
Kundu \emph{et al.} \cite{kundu2022concurrent} & &   &   \Checkmark   &  &  $\mathbf{M}_{pl}$     &
 -  & - & - & - & - & - & - & - & - & - & - & - & 88.2 \\
BMD + SHOT++ \cite{qu2022bmd} & &   &   \Checkmark   &  &  $\mathbf{M}_{pl}+\mathbf{M}_{em}$     &
 96.9   & 87.8 & 90.1 & \textbf{91.3} & 97.8 & 97.8 & 90.6 & 84.4 & 96.9 & \textbf{94.3} & 90.9 &45.9 & \textbf{88.7} \\
ProxyMix \cite{ding2022proxymix} &\Checkmark &   &   \Checkmark   &  & 
$\mathbf{M}_{vdg}+\mathbf{M}_{pl}$     &
  95.4   & 81.7 & 87.2 & 79.9 & 95.6 & 96.8 & 92.1 & 85.1 & 93.4 & 90.3 & 89.1 &42.2 & 85.7 \\
UTR \cite{pei2022uncertainty}  & &   &   \Checkmark   &  & 
$\mathbf{M}_{pl}$     &
\textbf{98.0}   & \textbf{92.9} & 88.3 & 78.0 & 97.8 & 97.7 & 91.1 & 84.7 & 95.5 & 91.4 & 91.2 &41.1 & 87.3 \\
FAUST \cite{lee2022feature}   & &   &   \Checkmark   &  & 
$\mathbf{M}_{pl}+\mathbf{M}_{em}$     &
96.7   & 77.6 & 87.6 & 73.3 & 95.5 & 95.4 & \textbf{92.9} & 83.6 & 95.3 & 89.5 & 87.7 &46.9 & 85.2 \\
DAMC \cite{zong2022domain} & \Checkmark &   &      &  & 
$\mathbf{M}_{iaa}$     &
95.4   & 86.9 & 87.2 & 79.7 & 94.6 & 95.7 & 91.2 & 84.9 & 95.7 & 91.5 & 81.9 &41.7 & 86.0 \\
SCLM \cite{tang2022semantic} & & \Checkmark  &   \Checkmark   &  & 
$\mathbf{M}_{nc}+\mathbf{M}_{pl}$     &
97.1   & 90.7 & 85.6 & 62.0 & 97.3 & 94.6 & 81.8 & 84.3 & 93.6 & 92.8 & 88.0 &55.9 & 85.3 \\
Kundu \emph{et al.} + SHOT++ \cite{kundu2022balancing} & \Checkmark   &   & \Checkmark   &    &    $\mathbf{M}_{vdg} + \mathbf{M}_{pl}$      &
 -  & - & - & - & - & - & - & - & - & - & - & - & 87.8 \\

 \bottomrule
\end{tabular}}
\end{table*}

 \textbf{Domain-based reconstruction is commonly used in SFDA.} The intuition of SFDA methods behind domain-based reconstruction is quite clear: it aims to reconstruct a new source or target domain to replace the missing source domain for supervision. Among this category of methods, intra-domain adversarial alignment methods were the first batch methods, such as BAIT \cite{yang2020casting} and 3C-GAN \cite{li2020model}. However, their influence is unenviable, probably because the source-similar samples selected in the target domain contain a lot of noise and are not representative enough, which is also why D-MCD \cite{chu2022denoised} targets at the de-noising problem. The attention of virtual domain generation is on the rise, probably because virtual domain generation methods~\cite{ding2022proxymix,lee2022confidence,kundu2022balancing,tian2021vdm} can be closely combined with self-training methods that play the supervised role of the target domain, thus together further improving the effect, e.g., reaching the highest accuracy of 90.7\% (Kundu \emph{et al.} \cite{kundu2022balancing}) on the Office-31 dataset.

 \textbf{Neighborhood clustering seems to be effective in image-based information extraction.} Compared with image style transformation methods \cite{hou2020source}, neighbor clustering is obviously more effective. For example, SCLM \cite{tang2022semantic} achieves the highest accuracy on Cl$\rightarrow$Ar and Cl$\rightarrow$Rw transfer tasks on Office-Home. The reason may be that neighbor clustering can better preserve the underlying structural information in clipart images. Besides, AaD \cite{yang2022attracting} achieves the second highest classification accuracy of 88\% on VisDA using only one loss function, which can be used as a simple but strong baseline.

 \textbf{Self-attention is promising, especially transformer-based SFDA methods.} Although the self-attention-based SFDA methods are less used in image classification \cite{yang2021transformer,yang2021generalized} and mostly seen in semantic segmentation \cite{liu2021source,kothandaraman2021ss,kothandaraman2023salad}, the transformer-based method TransDA \cite{yang2021transformer} achieves the highest accuracy of 79.3\% on the Office-Home dataset, which is 4.8\% higher than the second highest method \cite{kundu2022balancing}. This somewhat suggests that encouraging models to turn their attention to the object region may be quite effective for reducing domain shift. However, TransDA \cite{yang2021transformer} merely injects the transformer into the convolutional network, and it may be an interesting topic to see how the transformer can be better combined with the SFDA setting.

\section{Application}
Domain adaptation aims to reduce the cost of labeling target domain data, while source-free domain adaptation goes a step further to preserve the privacy of source domain data. Therefore, source-free domain adaptation and unsupervised domain adaptation methods \cite{wilson2020survey,liu2022deep,wang2018deep} are highly overlapped in their application areas. Currently, most of SFDA methods are applied in the field of computer vision and natural language processing.

\subsection{Computer Vision}
Most SFDA methods focus on image classification, which is the fundamental task in computer vision. With the popularity of large-scale datasets like VisDA \cite{peng2017visda} and DomainNet \cite{leventidis2021domainnet}, the demand of the adaptation capability also increases. Image classification can also extend to various application scenarios such as real-world image dehazing \cite{yu2022source}, cross-scene hyperspectral image classification \cite{xu2022source}, and blind image quality assessment \cite{liu2022source}. Semantic segmentation methods \cite{liu2021source,bateson2022source,prabhu2021augco,kothandaraman2023salad,bateson2020source,yang2022source,ye2021source} have also emerged rapidly and have been widely used, including multi-organ segmentation \cite{hong2022source}, cross-modal segmentation \cite{liu2023memory,kondo2022source}, cross-device \cite{yang2022source}, cross-central domain segmentation \cite{ye2022alleviating}, multi-site and lifespan brain skull stripping \cite{li2022source}, and road segmentation \cite{shenaj2023learning,you2021domain}. Others applications include object detection \cite{xiong2022source,zhang2021source,saltori2020sf}, person re-identification \cite{ding2022source}, and video analysis \cite{xu2022source,huang2022relative} can also benefit from SFDA.

\subsection{Natural Language Processing}

The application of source-free domain adaption in natural language processing (NLP) is still relatively limited. Related settings and studies in NLP include continuous learning \cite{de2019episodic,sahoo2020unsupervised} and generalization capabilities of pre-trained models \cite{hendrycks2020pretrained}. Laparra \emph{et al.} designed the SemEval 2021 Task 10 dataset~\cite{laparra2021semeval} on two tasks, i.e., negation detection and time expression recognition. Su \emph{et al.} \cite{su2022comparison} extended self-training \cite{yarowsky1995unsupervised}, active learning \cite{su2021university} and data augmentation \cite{miao2020snippext} baselines to the source-free setting for systematic comparison.

\subsection{Other Related Problems}
Domain generalization (DG) \cite{zhou2021domain,muandet2013domain}, which makes the model work on previously unseen target domains, can be seen as a relevant setting for domain adaptation, and some source-free DG methods \cite{niu2022domain,frikha2021towards} have also emerged, which can be seen as a variant of the data-based SFDA approach. Source-free zero-shot domain adaptation \cite{yazdanpanah2022visual,zhangsource} is proposed to alleviate the requirement for streaming data with small batch size and class distribution in test-time domain adaptation~\cite{varsavsky2020test}. Ondrej~\cite{bohdal2022feed} \emph{et al.} investigated feedforward source-free domain adaption, which is back propagation-free, to further protect user privacy as well as reduce overhead. There are also methods that combine source-free domain adaptation setting with federated learning \cite{shenaj2023learning}, black box test \cite{peng2022toward}, or robust transfer \cite{agarwal2022unsupervised} to meet different practical scenarios.

\section{Future Research Direction}
Despite the rapid growth of research on source-free domain adaptation and the resulting evolution on methodology and performance, the problem settings and targets studied in SFDA are still somewhat limited and homogeneous. Specifically, the classification tasks dominate the research of SFDA, which largely outpaces the growth of SFDA on other computer vision tasks. Below we discuss where the research landscape of SFDA can be potentially extended. 

\subsection{Enrichment of weak research lines}
Although we have discussed numerous SFDA methods in this paper, most of them focus on pseudo-labeling. With the categorization in Fig.~\ref{fig:framework}, some kinds of methods are still less explored, such as perturbed domain supervision, neighborhood clustering, self-attention especially transformer-based self-attention mechanisms. Contrastive learning \cite{khosla2020supervised} and image style translation \cite{huang2018multimodal} are still constructed in a relatively homogeneous and simple way, and it is worth introducing some of the latest methods \cite{zhan2022bi,yeh2022decoupled} in their respective fields. In addition, entropy minimization, although widely used in SFDA, is mostly used directly as an auxiliary loss in the form of entropy loss or information loss, lacking further innovation. In general, SFDA still has abundant room for enrichment in terms of the methodology.

\subsection{Further theoretical support}
Ben-David \emph{et al.} \cite{ben2010theory} derived a general UDA theory that upper-bounds the expected target error based on the distribution divergence of source and target domains, the error of the model on the source domain and the ideal joint error. However, the domain divergence is hard to measure in the source-free setting. Most current SFDA methods are motivated by intuition and achieve empirical success on current datasets. Liang \emph{et al.} \cite{liang2020we} suggested information maximization as a means of achieving deterministic and decentralized outputs. While there have been SFDA theoretical analyses centered on pseudo-labeling \cite{chu2022denoised}, multi-source domains \cite{ahmed2021unsupervised,dong2021confident}, and model smoothness \cite{li2022jacobian}, these approaches are only suitable for particular methods. Therefore, theoretical support that is universally applicable to SFDA is highly beneficial.

\subsection{More Applications}
Regarding computer vision tasks, most of the current SFDA methods focus on image analysis, while video data contains significantly more spatial and temporal semantic information, which presents greater processing challenges. Consequently, there are only a few methods specifically designed for video analysis. Furthermore, unsupervised domain adaptation has found numerous applications in natural language processing \cite{rotman2019deep, desai2019adaptive, rocha2019comparative}, time-series data analysis \cite{shen2022temperature, wu2022weighted}, recommendation systems \cite{zhangmulti, su2022cross}, and geosciences \cite{lucas2020unsupervised, nasim2022seismic}. However, these applications have not yet been extended to source-free settings, indicating that SFDA may realize its potential in these areas.

\subsection{Comprehensive datasets and evaluation}
The majority of datasets currently used for SFDA evaluation have balanced categories and clean data, however, some SFDA research has highlighted the significant impact of unbalanced categories \cite{li2021imbalanced} and noise \cite{agarwal2022unsupervised} on model accuracy. Additionally, some existing datasets are limited in size and no longer present sufficient challenges to assess the capabilities of modern domain adaptation methods. For example, the W to D task on the Office-31 classification dataset \cite{saenko2010adapting} has already reached an accuracy of 100\%. Thus, more diverse and challenging datasets are necessary for advancing SFDA research. Furthermore, a more comprehensive evaluation scheme, including model robustness and overhead, would provide a more complete understanding of model characteristics and improve the applicability of SFDA in various scenarios, such as edge devices.

\subsection{Extended settings}
This survey mainly covers the single-source closed-set setting of source-free domain adaptation, which is also the most extensively studied case. Nevertheless, as illustrated in Fig.~\ref{fig:label_set}, domain adaptation can be categorized into partial DA, open-set DA, multi-source DA, and multi-task DA, based on the relationship between the source and target domains. Furthermore, depending on various practical situations, the SFDA setting can be combined with test-time DA \cite{varsavsky2020test}, federated DA \cite{peng2019federated}, and active DA \cite{su2020active}. In the future, SFDA research is expected to become more diverse as researchers explore different settings and scenarios.

\section{Conclusion}
Unsupervised Domain adaptation, a crucial subset of transfer learning, facilitates the transfer of knowledge acquired from one labeled domain to another unlabeled domain, thereby reducing the need for extensive annotation of neural networks. The source-free setting, which lacks access to source domain data, satisfies privacy and security requirements in real-world scenarios and has rapidly gained attention since its emergence. In this paper, we provide a comprehensive overview of existing source-free adaptation methods and present a unified categorization framework. We analyze the experimental outcomes of more than 30 representative SFDA methods on the three most popular datasets, namely Office-31, Office-home, and VisDA, and modularize each method to facilitate comparisons. Lastly, based on our analysis and the current state of SFDA research, we suggest potential research directions that could benefit this community.


\ifCLASSOPTIONcaptionsoff
  \newpage
\fi



%
\bibliographystyle{IEEEtran}
\bibliography{mybibfile}

\balance

%








\end{document}